\documentclass[lettersize,journal]{IEEEtran}
\usepackage{amsmath,amsfonts}
\usepackage{algorithmic}
\usepackage{array}
\usepackage[caption=false,font=normalsize,labelfont=sf,textfont=sf]{subfig}
\usepackage{booktabs}       % professional-quality tables
\usepackage{amsfonts}       % blackboard math symbols
\usepackage{nicefrac}       % compact symbols for 1/2, etc.
\usepackage{microtype}      % microtypography
\usepackage{xcolor}         % colors
\usepackage[permil]{overpic}
\usepackage{float}
\usepackage{algorithm}
\usepackage{algorithmic}
\usepackage{setspace}

\usepackage{cite}
\usepackage{textcomp}
\usepackage{stfloats}
\usepackage{url}
\usepackage{verbatim}
\usepackage{graphicx}
\def\BibTeX{{\rm B\kern-.05em{\sc i\kern-.025em b}\kern-.08em
    T\kern-.1667em\lower.7ex\hbox{E}\kern-.125emX}}
\usepackage{balance}
\begin{document}
\title{Single-view Neural Radiance Fields with Depth Teacher}
\author{Yurui Chen, Chun Gu, Feihu Zhang, Li Zhang
 \IEEEcompsocitemizethanks{
Li Zhang is the corresponding author. (e-mail:lizhangfd@fudan.edu.cn).
Yurui Chen, Chun Gu and Li Zhang are with the School of Data Science, Fudan University.
Feihu Zhang is with University of Oxford.
 }
 }

\markboth{IEEE TRANSACTIONS ON IMAGE PROCESSING, ~2023}
{Yurui Chen, Chun Gu,
\MakeLowercase{\textit{(et al.)}:
Single-view Neural Radiance Fields with Depth Teacher}}

\maketitle

\begin{abstract}
Neural Radiance Fields (NeRF) have been proposed for photorealistic novel view rendering. However, it requires many different views of one scene for training.  Moreover,  it has poor generalizations to new scenes and requires retraining or fine-tuning on each scene. 
In this paper, we develop a new NeRF model for novel view synthesis using only a single image as input. We propose to combine the (coarse) planar rendering and the (fine) volume rendering to achieve higher rendering quality and better generalizations. We also design a depth teacher net that predicts dense pseudo depth maps to supervise the joint rendering mechanism and boost the learning of consistent 3D geometry.
We evaluate our method on three challenging datasets. It outperforms state-of-the-art single-view NeRFs by achieving 5$\sim$20\% improvements in PSNR and reducing 20$\sim$50\% of the errors in the depth rendering. It also shows excellent generalization abilities to unseen data without the need to fine-tune on each new scene. 

 \end{abstract}


\section{Introduction}
\IEEEPARstart{T}{he} method Neural Radiance Fields (NeRF)~\cite{mildenhall2020nerf} is proposed for photorealistic novel view synthesis. Given many views of the scene, it creates implicit multi-view geometry and learns for view synthesis. However, it has poor generalizations to new scenes and requires retraining or fine-tuning on each scene. 
 
 Recent work~\cite{Yu_2021_CVPR,Trevithick_2021_ICCV} has explored the ways of using a single image to train NeRF. They introduce a convolutional feature encoder to learn the image representation which gives it some limited generalization abilities to unseen scenes.  But, without fine-tuning, these methods produce many floats and artifacts in rendering novel views. 
 
  Multi-Plane Images (MPI) representation that learns multiple RGB images from a single image is also used in \cite{Wu_2021_ICCV,Tucker_2020_CVPR,wu2022remote} for  novel view synthesis. However, MPI heavily relies on the qualities of the planar images and needs plenty of image planes to avoid blurs. There is no strong 3D geometry constraint and it fails in many complex scenes.
  
  MINE~\cite{Li_2021_ICCV2} introduces the volume rendering of NeRF into the MPI. It runs faster and produces better depth rendering quality compared with single-view NeRFs~\cite{Yu_2021_CVPR,Trevithick_2021_ICCV}. However, the rendering quality heavily relies on the number of image planes. It needs high-resolution 4D volumes to store the 4-channel  (RGB and volume density) image planes that cost a large amount of GPU memory in both training and prediction.

Besides, some methods focus on shape or object information for novel view rendering.  AutoRF~\cite{N2022AutoRF} learns 3D object radiance fields from single-view observations.  CodeNeRF~\cite{Jang_2021_ICCV} learns separated embedding to disentangle shape and texture. Sharf~\cite{rematasICML21} uses shape information as guidance for learning NeRF from a single image.

There are also some other strategies introduced for single-view NeRF rendering. For example, 
PVSeRF~\cite{DBLP:journals/corr/abs-2202-04879} proposes a joint pixel-, voxel- and surface-aligned NeRF. 
Pix2NeRF~\cite{2022Pix2NeRF} introduces pi-GAN~\cite{pigan} to NeRF for learning novel view synthesis using a single image as input.

 In this paper, we propose a joint rendering mechanism that takes the MPI strategy for coarse sampling proposals and the MLP\&volume-based rendering~\cite{mildenhall2020nerf} for fine sampling and rendering. Then, both the coarse point samples and the fine samples are combined according to their geometry distribution to realize a more accurate joint rendering. More importantly, we introduce a depth teacher net that serves as the guidance for the joint rendering. The monocular depth teacher predicts dense pseudo depth maps that assist the consistent 3D geometry learning between the MPI, the fine volume, and the joint rendering. It also boosts the multi-view geometry consistency between the source view and the target novel views that 
helps handle the occlusions, reduce the blurs and floats, and render accurate depths. 
 
In the experiments,  we verify the effectiveness of our method on three challenging real-scene datasets (RealEstate10K~\cite{zhou2018stereo}, NYU~\cite{silberman2012indoor} and  NeRF-LLFF~\cite{mildenhall2020nerf}) for novel view synthesis or depth estimation. Given a single image as input, our method is shown able to produce higher qualities in both the RGB image rendering and depth map prediction. It far outperforms state-of-the-art methods~\cite{Li_2021_ICCV2,Yu_2021_CVPR} with improvements of 5$\sim$20\% in PSNR and SSIM for the RGB rendering and reduces 20$\sim$50\% of the errors for the depth prediction.

\section{Preliminary} \label{sec:preliminary}
\paragraph{Volume Rendering}
NeRF~\cite{mildenhall2020nerf} represents a scene as a continuous radiance field.
It takes the 3D position $\rm x_i\in \mathbb{R}^3$ and the viewing direction as input and outputs the corresponding color $\mathbf{c}_i$ with its differential density $\sigma_i$. 
NeRFs use volume rendering to render image pixels. For each 3D point $x_i$ in the space, its color can be rendered through the camera ray $\mathbf{r}(t) = \mathbf{o} + t\mathbf{d}$ with $N$ stratified sampled  bins between the near and far bounds of the distance.
\begin{small}
\begin{equation}
    \hat{\mathbf I}(\textrm{r}) = \sum_{i=1}^NT_i(1-\exp({-\sigma_i\delta_i})) \mathbf c_i,
    \\ \quad T_i = \exp\left( -\sum_{j=1}^{i-1}\sigma_j\delta_j \right). \label{eq:volume_rendering}
% \end{small}
\end{equation}
\end{small}
Where $\mathbf{o}$ is the origin of the ray, $T_i$ is the accumulated transmittance along the ray, $\mathbf c_i$ and $\sigma_i$ are the corresponding color and density at the sampled point $t_i$. $\delta_j = t_{j+1}-t_j$ refers to the distance between the adjacent point samples. 

Similarly, the depth map can be rendered as:
\begin{equation} \label{eq_render_depth}
    \hat{\mathcal{D}} = \sum_{i=1}^{N}T_i\big(1-\exp(-\sigma_{i}\delta_{i})\big) z_i.
    % \vspace{-3mm}
\end{equation}

\paragraph{Planar Neural Radiance Field}
Planar Neural Radiance Field, introduced by \cite{Li_2021_ICCV2}, is a perspective geometry for representing the camera frustum. For any pixel located on image coordinate $(x, y)$ , it represents the 3D location with candidate depth $z$ as $(x, y, z)$. Then, it learns $D$ multi-plane images (in a shape of $depth (D) \times height (H)\times width (W)\times 4$) with ($\mathbf{c}_i, \sigma_i$) at each 3D location $(x,y, z)$ to represent the image.

For novel view rendering, the intersection locations between the camera ray and each plane are sampled to achieve  $\mathbf{c}_i$ and $\sigma_i$ for image rendering using Eq. \eqref{eq:volume_rendering}.  The number of the sampled point $t_i$ equals the number of planes.

The planar rendering is faster than volume rendering and has better generalization abilities to unseen images. However, it's costly in terms of memory consumption since it requires storing a 4D volume for plane sampling. Also, the rendering quality heavily relies on the number of image planes and is thus not suitable for fine rendering.

\section{Method} \label{sec:method}

\begin{figure*}[t]
% \vspace{-15mm}
\centering
\begin{overpic}[width=1.0\textwidth]{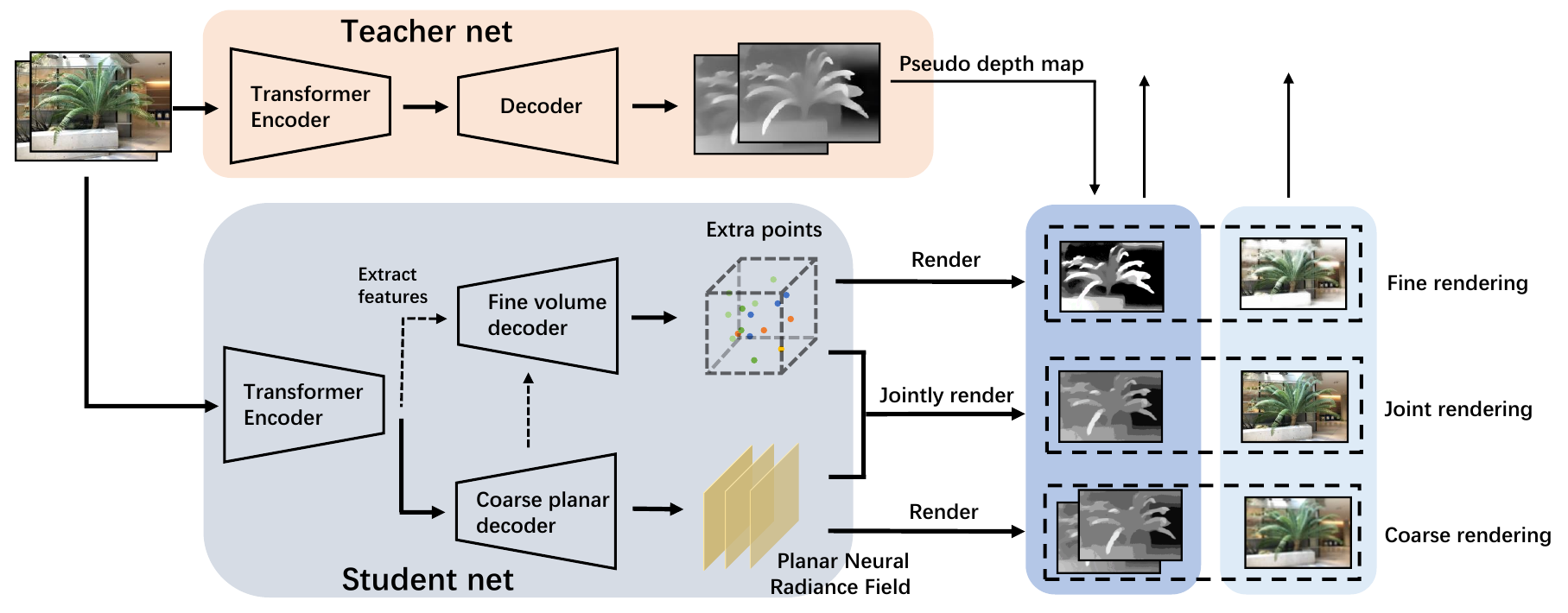}
\put(694,365){\footnotesize $\mathcal{L}_p, \mathcal{L}'_d$}%
\put(779,365){\footnotesize $\mathcal{L}_{\mathrm{L}1}, \mathcal{L}_{\mathrm{ssim}}$}%
\end{overpic}
% \vspace{-3mm}
   \caption{Our model contains two parts: the teacher net and the student net. Student net is responsible for novel view synthesis task. It takes a single RGB image as input and outputs a coarse planar radiance field which is later refined by the extra points predicted by the fine decoder. We then combine coarse sampling and fine sampling to jointly render the high-quality novel views and depth maps. The teacher net aims at supervising the student net on depth estimation and boosting geometry consistency.
   }
   \label{fig:pipline}
 \vspace{-2mm}
\end{figure*}
In this section, we describe our DT-NeRF for single-view-based novel view synthesis.
As illustrated in Figure \ref{fig:pipline}, 
our method consists of two major parts: 1) the depth teacher net, % (Sec. \ref{sec:teachernet}) 
that is based on the monocular depth estimation and predicts pseudo dense depth to supervise the student net rendering.
2) the student rendering net. % (Sec. \ref{sec:studentnet}).
It consists of a coarse planar rendering decoder %(Sec. \ref{sec:coarseplannar}) 
and a fine volume rendering decoder. % (Sec. \ref{sec:finevolume}). 
We use the depth teacher to supervise the joint rendering and boost the geometry consistency between the two renderings.

\subsection{Feature Encoder}
In order to generalize to new scenes without retraining or fine-tuning, we introduce a feature extractor to encode the input image. This allows the network to be trained across multiple scenes to learn a scene prior to generalizations to unseen data.
The whole output image feature tensor is then used as the input of the coarse planar decoder to learn 4-channel (RGB-$\sigma$) MPI, while the sampled point features are fed to the fine volume decoder for fine volume rendering.

We use a vision transformer~\cite{Ranftl_2021_ICCV} as our feature encoder. The vision transformer contains a ResNet50 backbone and 12 transformer blocks. %Following , %we extract the features from the forward process of vision transformer as the output of the encoder. 
In order to reduce the size of the model, the transformer encoder can be shared by the teacher net and the student net.

\subsection{Depth Teacher}

Previous methods~\cite{Li_2021_ICCV2,Tucker_2020_CVPR} are basically supervised by color information (and some point clouds), their depth estimation is not always reliable, resulting in blurred boundaries or unnatural holes.
To generate a more reasonable depth estimation and boost the 3D geometry consistency between the novel views and the input source image, we add a depth teacher net $\mathcal{G}$ to generate a dense pseudo depth map from a single input RGB image. The dense depth maps of the source view and target view generated from $\mathcal{G}$ are used as pseudo labels to supervise the student net for both RGB and depth rendering.

Our teacher net is adapted from \cite{Ranftl_2021_ICCV} which consists of a transformer encoder and a convolutional depth prediction decoder. The dense depth map generated by the teacher net builds strong alignment between RGB labels and depth pseudo labels. It contains more 3D prior knowledge than the previous method~\cite{Li_2021_ICCV2,Tucker_2020_CVPR}, and boosts the student net to learn more consistent geometries.

\subsection{Student Joint Rendering} \label{sec:studentnet}

\begin{figure}[t]
 \vspace{-5mm}
\begin{center}
    \includegraphics[width=0.99\linewidth]{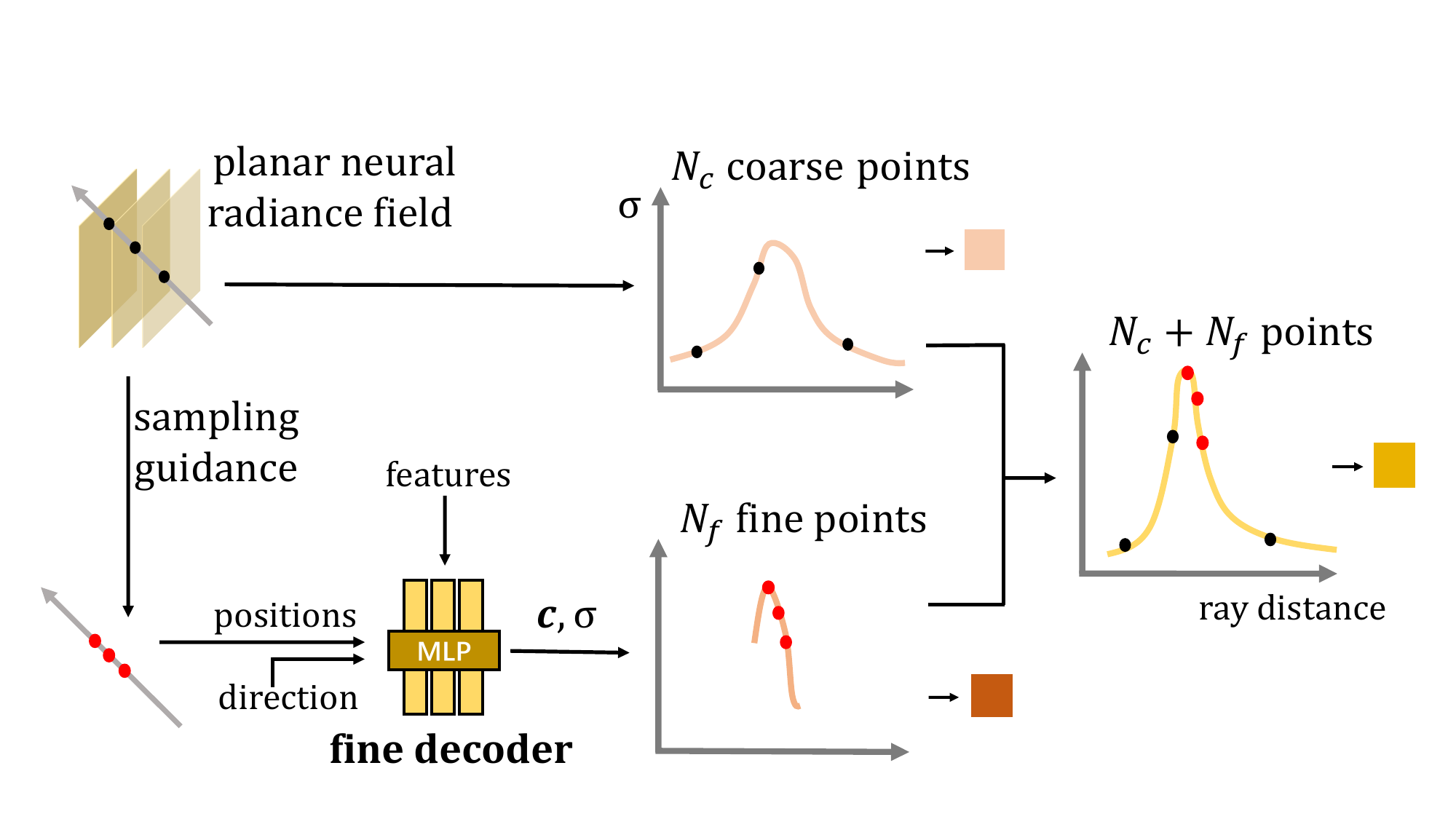}
\end{center}
 \vspace{-4.5mm}
   \caption{Joint sampling and rendering consists of the coarse planar rendering (top) and the fine volume rendering (bottom). For rays emitted from the target view camera, the model can roughly estimate the probability density function of the weight $T_i$ from $N_c$ multi-plane images. Then, additional $N_f$  point samples are selected by the importance sampling. The corresponding point features are extracted from the  output of the feature encoder, and then input to the fine decoder together with positions and directions to obtain the RGB-$\sigma$. Finally, $N_c+ N_f$ points are combined for a joint rendering.}
   \label{fig:arch}
 %\vspace{-2mm}
\end{figure}

The teacher net outputs pseudo dense depth maps to supervise our student joint rendering networks. The student net consists of two components: 1) the coarse planar NeRF that learns coarse planar sampling and rendering as guidance to resample 2) the fine volume rendering. Finally both the samples are employed for joint rendering with Eq. \eqref{eq:volume_rendering} (illustrated in Figure \ref{fig:arch}). This is different from the coarse and fine sampling of the original NeRF~\cite{mildenhall2020nerf} where only fine samples are used for final RGB and depth rendering.

\paragraph{Coarse Planar Neural Radiance Field} \label{sec:coarsplannar}
Our planar neural radiance field network consists of an encoder and a decoder. The feature encoder takes a source view image as input and outputs a feature tensor. The feature tensor is input to the coarse planar decoder and predicts a 4-channel image ($\mathbf{c}_i,\sigma_i$) at each candidate depth $z_i$ (total $N_c$ candidate depth values). %in the shape of 

Similar to \cite{Li_2021_ICCV2}, for each pixel $(x_t, y_t)$ in the target novel views, we warp it to the source multi-plane images according to the camera rays. 

We define the homography transformation $\mathcal{W}(\cdot)$ between the pixel coordinates of the target novel view $(x_t,y_t)$ and the multi-plane images of the source view $(x_i, y_i, z_i)$. %:
For simplicity, we use $(x_i,y_i, z_i)=\mathcal{W}_{i}(x_t,y_t)$ to denote the mapping to a plane with depth of $z_i$ at the source camera (More details can be found in \cite{Li_2021_ICCV2} or the supplementary). 
$N_c$ coarse samples $(\mathbf{c}_i, \sigma_i)$ are selected at the warped location $(x_i, y_i, z_i)$ in $N_c$ image planes.  We then use Eq. \eqref{eq:volume_rendering} to render the coarse images.

\paragraph{Fine Volume Rendering}\label{sec:finevolume}

The planar rendering runs faster and has a better generalization to unseen data. However, it costs a large amount of memory to store multi-plane images and render high-resolution novel views. It's necessary to use a  small $N_c$  to reduce the memory costs. % and propose to implement a different fine volume rendering ways to further render more image details. 
However, insufficient sampling from the image planes (a small $N_c$) usually leads to blurry novel views. Also, the coarse planar rendering doesn't take the ray direction into account and can't model complex view-dependent effects (e.g. lighting variations).  %method which means $c$ and $\sigma$ are just functions of coordinates. 

We build a fine volume MLP decoder to boost the rendering and select more important RGB-$\sigma$ values of interested points in the space. %We improve the rendering by sampling new important points from the camera rays. 
The new importance sampling is guided by the coarse planar sampling results. Since the volume rendering of Eq.~\eqref{eq:volume_rendering} can be interpreted as a weighted sum of all sampled colors $\mathbf{c}_i$, we compute the weights of colors $T_{i}$ after the rendering with the coarse planar neural field for each ray. We can then obtain a probability density function (PDF) estimation of the weight along the ray by normalizing these weights  $\hat{T}_{i}=T_{i} / \sum_{j} T_{j}$. Then, we sample a new set of (a total of $N_f$) positions $\mathbf{x}_i = (x_i, y_i ,z_i)$ from this distribution using inverse transform sampling~\cite{mildenhall2020nerf}. 
%Fine decoder, 

According to the sampled 3D positions, we extract features from multi-plane images. Moreover, we project the 3D spatial coordinate $\mathbf{x}$ to the source image plane using the projection matrix $\mathbf{P}\sim [\mathbf{R}, \mathbf{T}, \mathbf{K}]$ ($(x_s, y_s) = \mathbf{P}\mathbf{x}$, where $\mathbf{R},\mathbf{T}$ are the camera rotation and translation from the target view to the source view and $\mathbf{K}$ is the camera intrinsic parameters).

Given an input feature tensor (output by the feature extractor) $\mathbf{F}_{src}$,  we use the projected pixel coordinates $(x_s, y_s)$ to extract the new feature samples $\mathbf{f}$. 
Then, the sampled feature vectors $\mathbf{f}_i$ together with the position and view direction are fed to the fine decoder network (implemented by MLP layers) and output the color $\mathbf{c}$ and density $\sigma$. This is similar to the original NeRF~\cite{mildenhall2020nerf}.

\paragraph{Joint Sampling and Rendering}
%Both 
We combine the $N_c$ coarse samples and the $N_f$ fine samples by placing them correctly along the camera rays (as illustrated in Figure \ref{fig:arch}). The joint rendering takes the two types of $\mathbf{c}$ and $\sigma$ as inputs to the rendering function of Eq. \eqref{eq:volume_rendering} and outputs the better RGB and depth rendering results.

Since the volume rendering and the coarse planar rendering are not in the same space.  To correctly place the coarse and fine point samples along the camera rays (as illustrated in Figure \ref{fig:arch}).  It is important to preserve a consistent 3D geometry between the coarse MPI and the fine volume. We propose to use the dense pseudo depth maps predicted by our depth teacher to supervise the joint rendering and make them consistent with each other.

\section{Implementation}

In order to accelerate the convergence in training, we divide our training into two stages. We first fine-tune the teacher net, then, train our student net with the depth teacher fixed.

\subsection{Depth Teacher Pre-training}
Since the depth map predicted by the teacher net is not completely accurate and there may be problems with scale, we pre-train/fine-tune the teacher net using the available sparse point cloud before training student net. Following \cite{monodepth2}, the pre-training is supervised by the projection color errors, the gradient consistency and the sparse 3d points (achieved by SfM algorithm~\cite{colmap}).
\begin{align}
     \mathcal{L}^\mathcal{G} = \mathcal{L}_{proj} + \mathcal{L}_{smooth} + \mathcal{L}_p \label{eq:fine-tune loss},
\end{align}
where the reprojection error $\mathcal{L}_{proj}$ and the edge-aware smoothness loss $\mathcal{L}_{smooth}$ are the same as those in \cite{monodepth2}, and $\mathcal{L}_p$ denotes the point cloud loss (Eq. \eqref{eq:point_loss}).

\subsection{Learning Joint Rendering}

Our student net mainly has three parts: 1) the ``coarse'' view rendered by planar neural field, 2) the ``fine'' view rendered by fine volume rendering with only importance sampling points, and 3) the ``joint'' rendering results  that combined both the coarse samples and the fine samples. Each part will output the corresponding RGB image $\hat{\mathrm{I}}$ and depth map estimation $\hat{\mathcal{D}}$ respectively. The joint rendering is supervised by 
\begin{equation}
    \mathcal{L} = \mathcal{L}_c +  0.4\mathcal{L}_f + \mathcal{L}_j.
\end{equation}

Where $\{c,f,j\}$ represents the ``coarse'', ``fine'' and ``joint'' rendering. Each part has the same loss that consists of $L1$ loss $\mathcal{L}_{\mathrm{L}1}$,  $SSIM$ loss~\cite{wang2004image} $\mathcal{L}_{\mathrm{ssim}}$, sparse point cloud loss $\mathcal{L}_{p}$ (if available) and pseudo depth loss $\mathcal{L}'_{d}$:
\begin{equation}
\mathcal{L}_{c/f/j}= \mathcal{L}_{\mathrm{L}1}+\lambda_{\mathrm{ssim}} \mathcal{L}_{\mathrm{ssim}}+\lambda_{p} \mathcal{L}_{p}
+ \lambda'_{d} \mathcal{L}'_{d}.
\end{equation}

\begin{equation}
\mathcal{L}_{\mathrm{L}1}=\left|\hat{\mathrm{I}}-\mathrm{I}\right|,  \mathcal{L}_{\mathrm{ssim}}=1-\operatorname{SSIM}\left(\hat{\mathrm{I}}, \mathrm{I}\right)
\end{equation}
The RGB $L1$ loss and $SSIM$ loss make the target novel view $\hat{\mathrm{I}}$ generated by our model match the ground truth image $\mathrm{I}$.

%\paragraph{Sparse point cloud loss}
If there are some sparse point clouds $\mathbf{P}$ available (usually generated by SfM method~\cite{colmap}), the  sparse point cloud loss can be used as

\begin{small}
\begin{equation}
\mathcal{L}_{p} =\frac{1}{\left|\mathbf{P}\right|} \sum_{(x, y, z) \in \mathbf{P}}\left(\ln \frac{\hat{\mathcal{D}}(x, y)}{s}-\ln \frac{1}{z}\right),
\label{eq:point_loss}
\end{equation}
\end{small}
where, $\hat{\mathcal{D}}$ represents the rendered disparity  (inverse depth) map and $s$ represents the scale factor relative to the point clouds set $\mathbf{P}$. The point cloud loss is applied for both the source view  and the target novel view rendering.

%\paragraph{Scale Calibration}
Since the scale $s$ between the disparity map generated by depth teacher network, student net and the point cloud  are usually different, it is necessary to scale them to a uniform scale before supervising training. When point cloud $\mathbf{P}$ is available, following \cite{Li_2021_ICCV2}, we unify the point cloud scale by
\begin{small}
\begin{equation}
s=\exp \left[\frac{1}{\left|\mathbf{P}\right|} 
\sum_{(x, y, z) \in \mathbf{P}}\left(\ln \frac{1}{z}-\ln \hat{\mathcal{D}}(x, y) \right)\right].
\end{equation}
\end{small}

For the pseudo depth loss $L'_d$, the $L1$ depth loss and the gradient  loss are used as

\begin{align}
\begin{array}{rrl}
    \mathcal{L}'_d &=& |\hat{\mathcal{D}}-\mathcal{D}^* |
    + \lambda_{grad}( |\partial_{x}(\hat{\mathcal{D}})-\partial_{x}(\mathcal{D}^*)| \\&+& |\partial_{y}(\hat{\mathcal{D}})-\partial_{y}(\mathcal{D}^*)|).
    \end{array}
    \label{eq:loss_dpt_l1}
\end{align}

$\hat{\mathcal{D}}$ and $\mathcal{D}^*$ are the scaled disparity  (inverse depth) maps predicted from the student net and the teacher net respectively. $\partial_{x}$ and $\partial_{y}$ are gradients of the disparity maps. We apply these two losses in both the source and the target view to boost the geometry consistency between the student rendering net and the depth teacher net.

In practice, for stable training and faster convergence, we train our student net by two steps. We first train the coarse planar neural radiance field, then fix the coarse planar rendering, and set loss to $0.4\mathcal{L}_f+\mathcal{L}_j$ to train fine decoder.

\subsection{Inpainting Refinement}
Since it's difficult to render image borders of the novel views when it is out of the field of view of the source image, we implement a light-weighted inpainting module to refine both the RGB and the depth results at the occlusions and image borders.  We leverage the predicted depth map of the source view to compute the occlusion mask of the target view by depth warping. We then follow \cite{suvorov2021resolution} to learn the inpainting, but change to use a four-channel (RGB-D) representation for both the input and the output. Inspired by \cite{han2022single}, we use the warp-back strategy to augment the data for training the inpainting module.

\subsection{Training Details}

For our experiments, $N_c$ is fixed to 32, $N_f$ is 16, $\lambda_{\mathrm{ssim}}$, $\lambda_{L1}$ and $\lambda_{p}$  are set to 1. We use the
Adam Optimizer~\cite{kingma2014adam} with an initial learning rate of 0.001 for both the planar radiance decoder and fine decoder, 1e-5 for transformer encoder, and 1e-4 for depth teacher $\mathcal{G}$'s decoder. Our fine decoder is a light-weighted MLP module which has 5 hidden layers with 64 channels. Fine decoder processes a 150-dimension feature and output 4-channel $(\textbf{c}, \sigma)$ prediction. Before feeding positions into the fine decoder, we apply a positional encoding~\cite{mildenhall2020nerf} which maps 3D coordinates into a 63-dimension feature space.

\section{Experimental Results}

\subsection{Datasets}
We use NeRF-LLFF~\cite{mildenhall2020nerf} and Realestate10k~\cite{zhou2018stereo}  datasets for training and validation. Besides, we also use NYUv2 dataset~\cite{silberman2012indoor}  as the test set to evaluate the depth rendering. 
\paragraph{NeRF-LLFF}
NeRF-LLFF~\cite{mildenhall2020nerf} consists of 8 scenes. Following \cite{Li_2021_ICCV2}, we select images in each scene as the test set which consists of 35 images, and the rest 270 images are used as the training set.
In experiments on NeRF-LLFF, %$D$ is fixed to 32, $N_f$ is 16, 
we set $\lambda'_{d}=10$, $\lambda_{grad}=5$.
The resolution is set to 512$\times$384. We fine-tune our depth teacher for 1,000 iterations, train coarse planar rendering for 20,000 iterations and another 10,000 iterations for fine rendering. We use a batch size of 4. 
The learning rate decays every 8000 steps.

\paragraph{RealEstate10K}
RealEstate10K~\cite{zhou2018stereo} is a large dataset that consists of more than 70,000 video sequences. Limited by the available computing resource, we randomly choose 1,000 sequences from the pre-split training set to train our model and test it on 600 randomly selected sequences from the test set. %Each video sequence is  
For training and testing, we sample the source and target view pairs at a 10-frame interval from the video sequences, which gives us 32,000 training pairs (of source and target views) and 2,400 test pairs. %For training in RealEstate10k, we set 
In experiments on Realestate10k dataset, %$D$ and $N_f$ is 32 and 16,  
we set $\lambda'_{d}=1$, $\lambda_{grad}=20$.
The input resolution is set to 384$\times$256. We fine-tune our $\mathcal{G}$ 2,000 iterations, train the coarse planar rendering for 10,000 steps with a batch size of 24 and another 20,000 iterations for the fine decoder with a batch size of 8. 
The learning rate decays at 1000 steps and 8000 steps.

\begin{figure*}[th!] \centering
\setlength{\abovecaptionskip}{4pt}
\setlength{\belowcaptionskip}{-8pt}
\small
\setlength{\tabcolsep}{1pt}
\resizebox{1.0\textwidth}{!}{
\begin{tabular}{ccccc}
     RGB & MINE & PixelNeRF & Ours (w/o $\mathcal{L}'_d$) & Ours \\
     \includegraphics[width=0.19\textwidth]{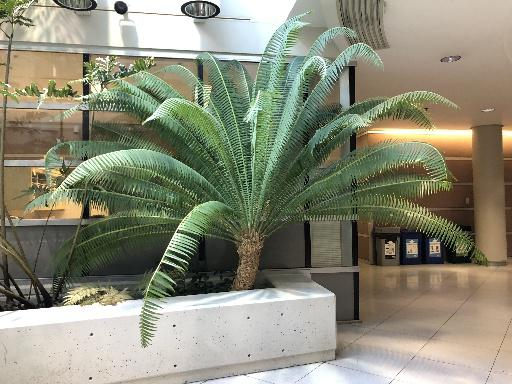} & 
     \includegraphics[width=0.19\textwidth]{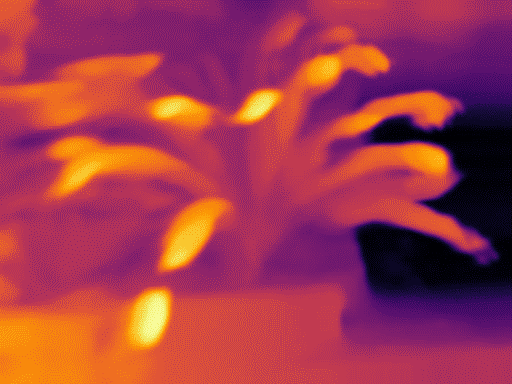} &
     \includegraphics[width=0.19\textwidth]{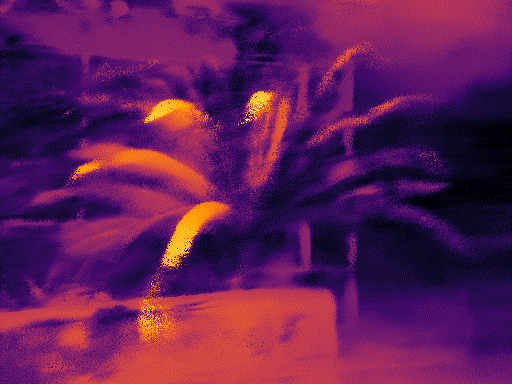}& 
     \includegraphics[width=0.19\textwidth]{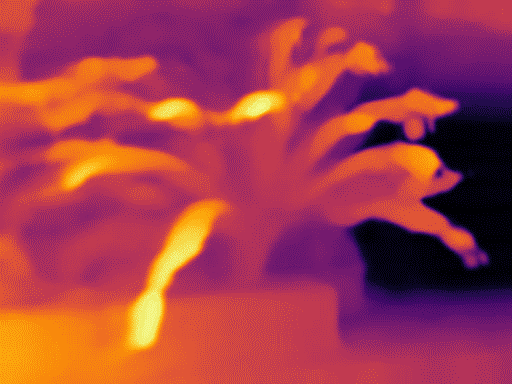} &
     \includegraphics[width=0.19\textwidth]{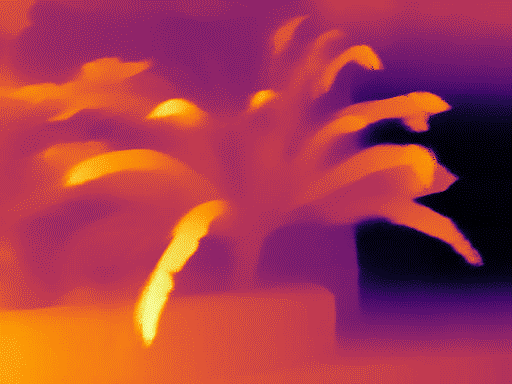} 
     \\
     \includegraphics[width=0.19\textwidth]{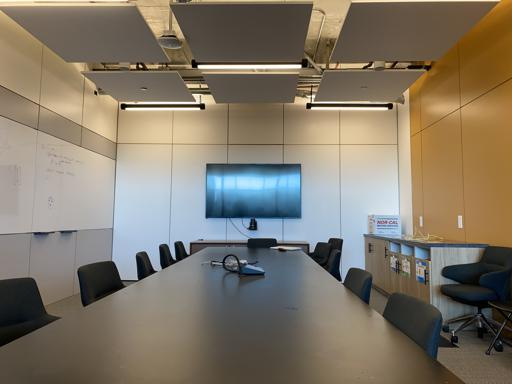} &
     \includegraphics[width=0.19\textwidth]{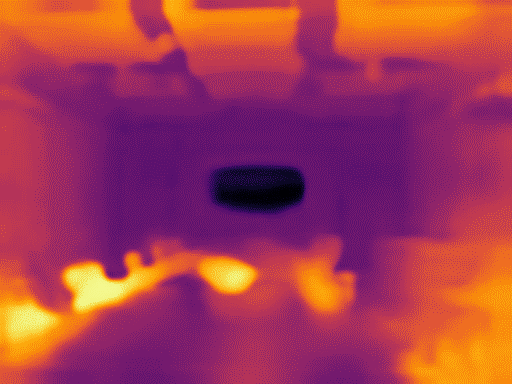} &
     \includegraphics[width=0.19\textwidth]{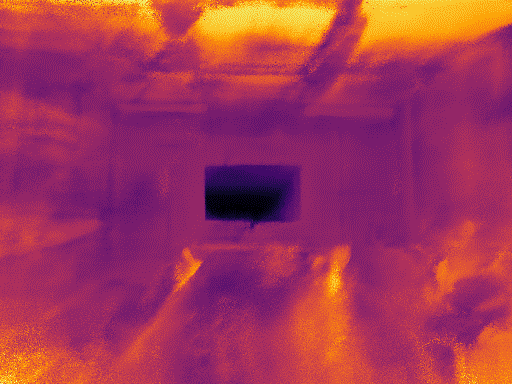}& 
     \includegraphics[width=0.19\textwidth]{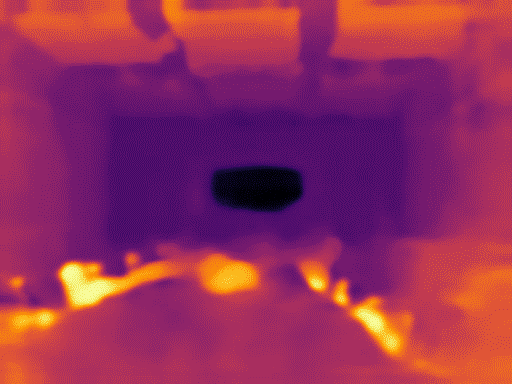} &
     \includegraphics[width=0.19\textwidth]{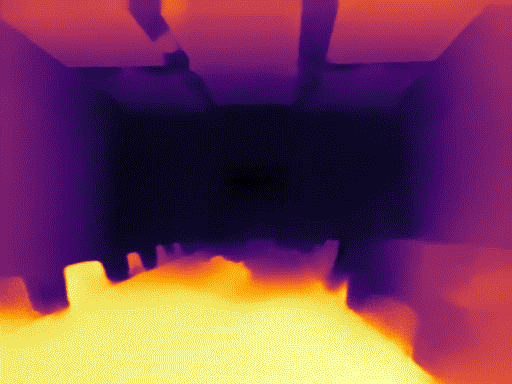} \\
\end{tabular}
}
    \caption{Effects of the pseudo depth loss. Even without the depth teacher, our method can achieve better depth maps compared with MINE and PixelNeRF. The depth teacher with pseudo depth loss further improves the quality of the depth rendering.} 
    % \vspace{-8pt}
    \label{fig:llff_depth}
\end{figure*} 

\subsection{Ablation Study}

\begin{figure*}[t]
\centering

%%%%%%%%%%%%%%%%%%
\subfloat[Planar rendering]{
\begin{minipage}[b]{0.48\linewidth}
     \includegraphics[width=0.99\linewidth]{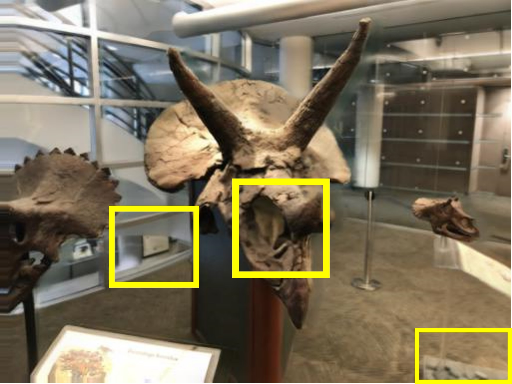}\\
     [0.0mm]\\
     \includegraphics[width=0.99\linewidth]{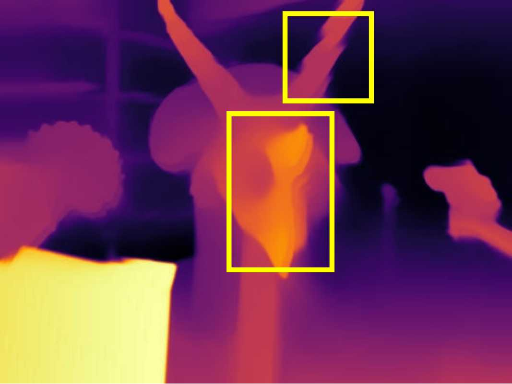} 
\end{minipage}
}
\hspace{0.2mm}
\subfloat[Planar + volume rendering]{
\begin{minipage}[b]{0.48\linewidth}
     \includegraphics[width=0.99\linewidth]{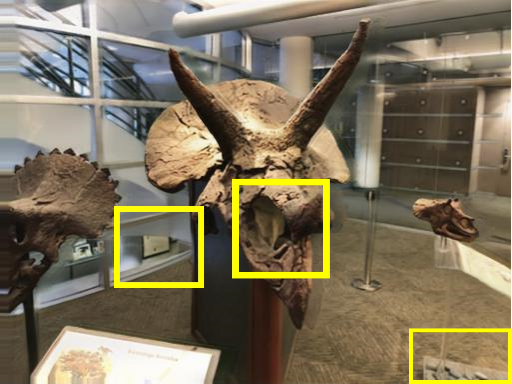}\\
     [0.0mm]\\
     \includegraphics[width=0.99\linewidth]{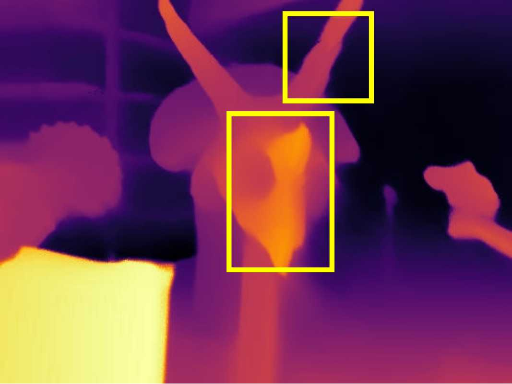}
\end{minipage}
}
%%%%%%%%%%%%%%%%%

%}
    % \vspace{1pt}
    \caption{Influence of fine volume rendering with importance sampling strategy. The improved areas are highlighted by yellow boxes. Compared with the pure planar rendering, extra fine point samples gives more image details and sharper depth edges.} 
    \label{fig: llff_is}
    % \vspace{-4pt}
\end{figure*}

As shown in Table \ref{tab:baseline_compare_detail}, we verify the effectiveness of different rendering components and study the effects of different settings on the LLFF dataset. We find that both our planar rendering and the volume rendering perform better than the baseline PixelNeRF. The joint rendering achieves a better PSNR and SSIM compared with each individual rendering way.
% add supp volume rendering
It should be noted that additional importance sampling can boost the planar rendering method for preserving more fine image/depth details. Since the number of planes sampled is limited, blur and artifacts are unavoidable. 
As shown in Figure \ref{fig: llff_is}, compared with the pure planar rendering~\cite{Li_2021_ICCV2},
the joint rendering predicts more details and fewer blurs in both the RGB and depth rendering. 
% Figure \ref{fig: llff_is} shows how extra fine points solve this problem and lead to a better rendering.
And the full setting with the depth teacher supervision achieves the best PSNR, SSIM and LPIPS.
Figure \ref{fig:llff_depth} also compares our method with the pure planar NeRF (MINE) and the pure volume rendering PixelNeRF. The joint rendering without our depth teacher achieves better depths in object edges and occlusions. By introducing the depth teacher supervision $\mathcal{L}'_d$, 
 the rendering quality of the  depth maps is significantly improved. 

Although teacher net (monocular depth estimation) is able to predict high-quality depth maps on the input image, it can't render depth maps for novel view. The  student net renders both the RGB and the depth maps for the novel views which are very important for applications like video editing, augmented reality etc. 
\paragraph{Coarse/Fine Point Sampling}
\begin{table}[t]
% \vspace{1mm}
\caption{Performances of our method using different $N_c$ and $N_f$ on NeRF-LLFF dataset. }
\centering
 \setlength{\abovecaptionskip}{4pt}
 \setlength{\belowcaptionskip}{-6pt}
% \small 
\begin{tabular}{cc|ccc}
\hline 
$N_c$ & $N_f$     & LPIPS$\downarrow$  & SSIM$\uparrow$  & PSNR$\uparrow$ \\
\hline
MINE (32) & -    & 0.386 & 0.531 & 18.20 \\ 
MINE (64) & -    & 0.424 & 0.539  & 18.13 \\ 
16 & 0     & 0.328 & 0.548 & 18.78 \\ 

16 & 16      & 0.396 & 0.585& 19.07 \\ 

16 & 32     & 0.347 & 0.608   &19.17 \\ 

32 & 0 & 0.305  & 0.612 & 19.41 \\ 

32 & 16    & 0.317  & 0.637 & 19.54 \\ 

32 & 32     & 0.292 & \textbf{0.650} & \textbf{19.57} \\ 

64 & 0   & 0.315 & 0.626   & 19.36 \\ 

64 & 16    & 0.293 & 0.641  & 19.38 \\ 

64 & 32 & \textbf{0.291}  & 0.642 & 19.39 \\

\hline
\end{tabular}
\label{tab:llff_ncnf_compare}
%\vspace{-4pt}
\end{table}
As shown in Table \ref{tab:llff_ncnf_compare}, we report the performance when using different coarse and fine samples. We find that using more  point samples (either coarse planar sampling $N_c$ or the fine volume samples $N_f$) will improve the rendering quality. When using 32 coarse planar samples and 32 fine volume samples, our method performs best. But, using many more planar samples (e.g. $N_c=64$) will not produce better results. %This might because  too many image planes are hard to learn. 

% It should be noted that additional importance sampling can boost the planar rendering method for preserving more fine image/depth details. Since the number of planes sampled is limited, blur and artifacts are unavoidable. Figure \ref{fig: llff_is} shows how extra fine points solve this problem and lead to a better rendering.

\begin{table}[t]
% \vspace{1mm}
\centering
\caption{Comparison with different encoder backbones. We test two different feature encoder backbones, the default transformer backbone VIT and the ResNet50 (used in MINE).  We set $N_c=32, N_f=16$. Our method performs far better than MINE when using the same backbone network (ResNet50). ViT slightly improves the PSNR on LLFF dataset when compared with ResNet50, and there are no significantly improvements on RealEstate10k dataset.}
% \small 
\begin{tabular}{ccc|c}
\hline 
Method&Backbone & Dataset   & PSNR$\uparrow$ \\
\hline
MINE & ResNet50 & LLFF  & 18.2\\
Ours&ResNet50 & LLFF&  19.3 \\ 
Ours&ViT & LLFF &  \textbf{19.5} \\ 
\hline
MINE & ResNet50 & RealEstate(small)  & 24.6\\
Ours&ResNet50 & RealEstate(small)& 25.0 \\ 
Ours&ViT & RealEstate(small) & \textbf{25.0} \\ 
\hline
\end{tabular}
\label{tab:llff_backbone_compare}
%\vspace{-4pt}
\end{table}

\paragraph{Different Backbones}
We test two different feature encoder backbones, the default transformer backbone ViT and the ResNet50 (used in MINE). When using the same ResNet50 backbone, our method achieves 19.3 in PSNR which is 6\% improvement compared with MINE under the same setting. 
ViT slightly improves the PSNR on LLFF dataset when compared with ResNet50, and there is no improvement on RealEstate10K dataset.  The improvements are mainly from our joint rendering and depth teacher guidance strategies.

\begin{figure}[t] \centering
%\small
\setlength{\abovecaptionskip}{4pt}
\setlength{\belowcaptionskip}{-5pt}
\setlength{\tabcolsep}{1pt}
\resizebox{1\linewidth}{!}{
\begin{tabular}{ccc}
      PixelNeRF & Ours (only volume) & GT\\ 

     \includegraphics[width=0.20\textwidth]{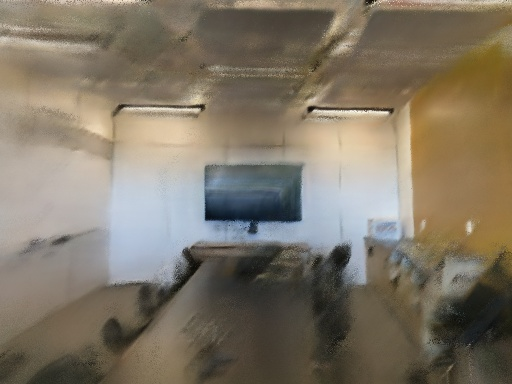} & 
     \includegraphics[width=0.20\textwidth]{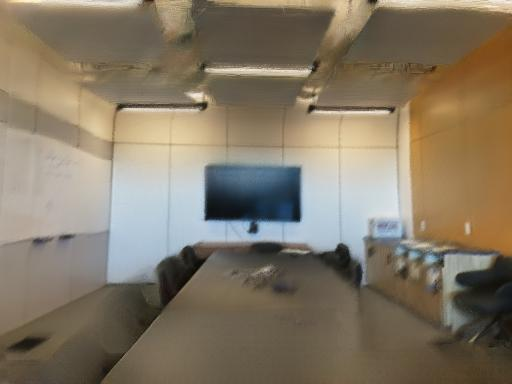}
     &
   \includegraphics[width=0.20\textwidth]{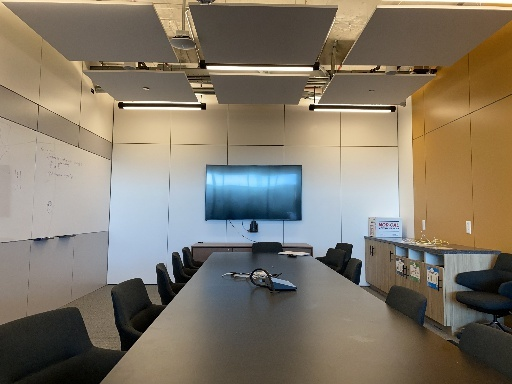} 
     \\
\end{tabular}
}
    \caption{Comparisons with PixelNeRF using just our volume rendering. With just 16 points selected, our volume rendering are far better than PixelNeRF (96 point samples).}
   % \feihu{add a new last column with joint rendering.}} 
    % \vspace{-8pt}
    \label{fig:llff pixelnerf}
\end{figure} 

\begin{figure*}[th!] \centering
\setlength{\abovecaptionskip}{6pt}
\setlength{\belowcaptionskip}{-8pt}
\small
\setlength{\tabcolsep}{1pt}
\resizebox{1.0\textwidth}{!}{
\begin{tabular}{cccccc}
      Input &MPI & MINE & PixelNeRF & Ours & GT
     \\ 
     \includegraphics[width=0.16\textwidth]{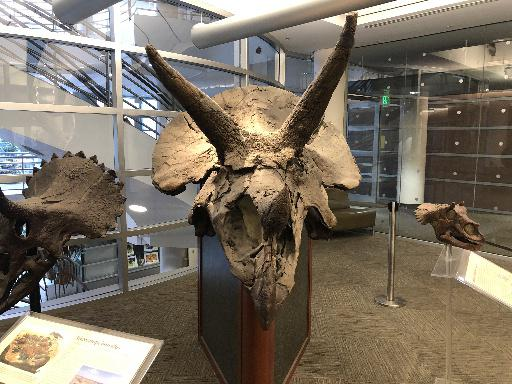}  & \includegraphics[width=0.16\textwidth]{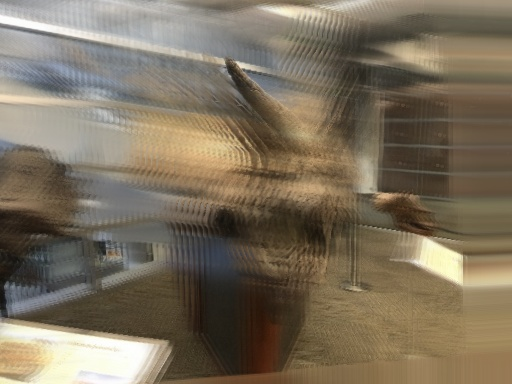} & 
     \includegraphics[width=0.16\textwidth]{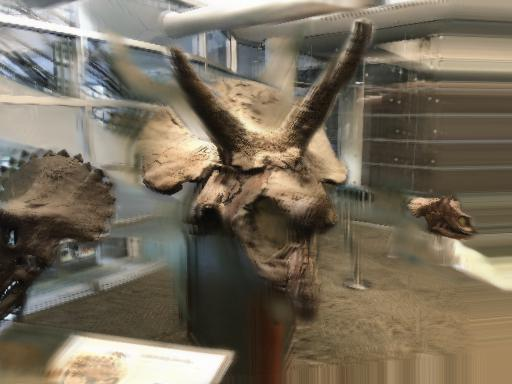}& 
     \includegraphics[width=0.16\textwidth]{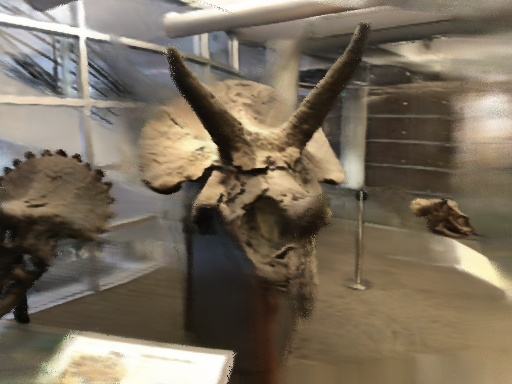}& 
     \includegraphics[width=0.16\textwidth]{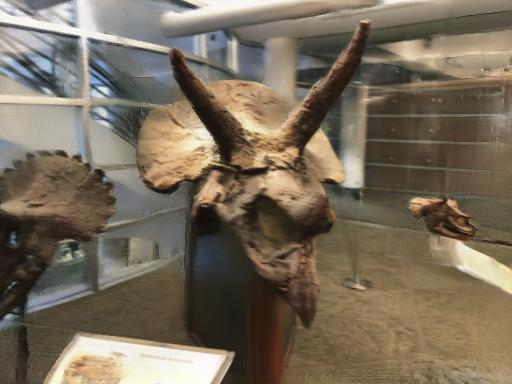} & \includegraphics[width=0.16\textwidth]{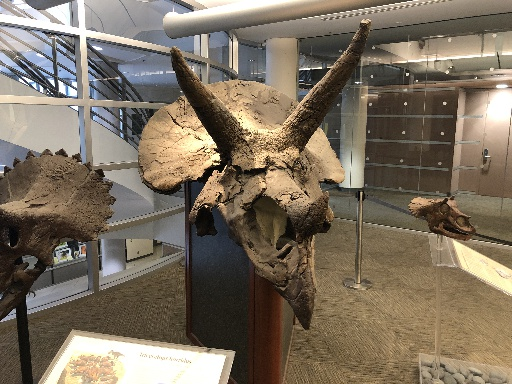}
    \\
    \includegraphics[width=0.16\textwidth]{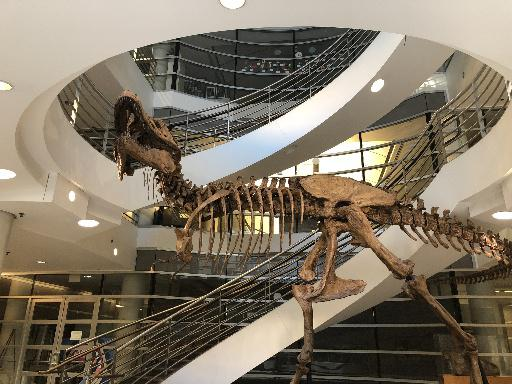}  & \includegraphics[width=0.16\textwidth]{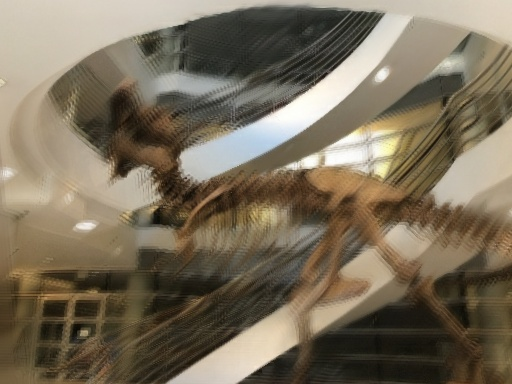} & 
    \includegraphics[width=0.16\textwidth]{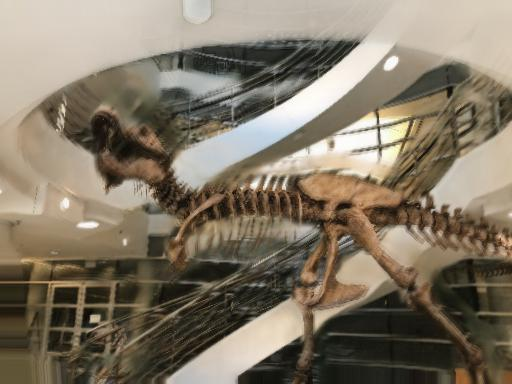}& 
    \includegraphics[width=0.16\textwidth]{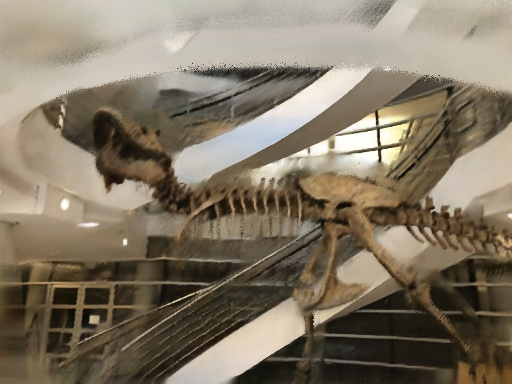}& 
    \includegraphics[width=0.16\textwidth]{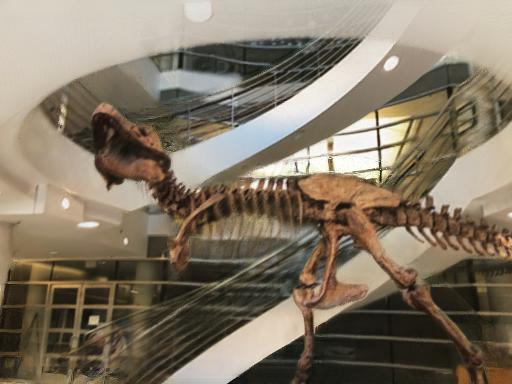} & \includegraphics[width=0.16\textwidth]{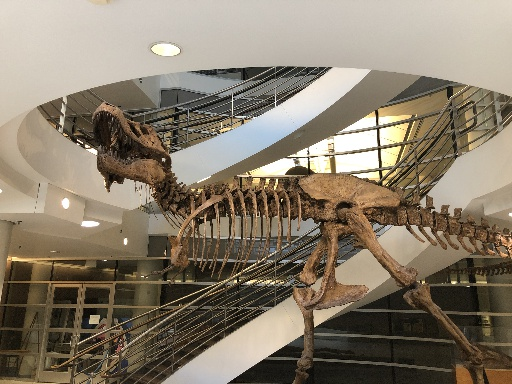}
     \\
     \includegraphics[width=0.16\textwidth]{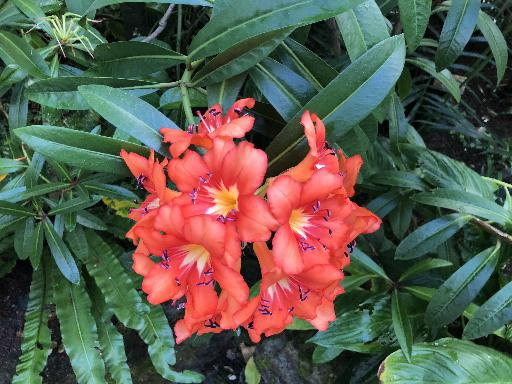}  & \includegraphics[width=0.16\textwidth]{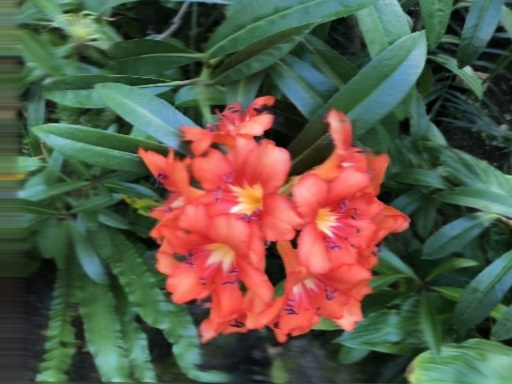} 
     & \includegraphics[width=0.16\textwidth]{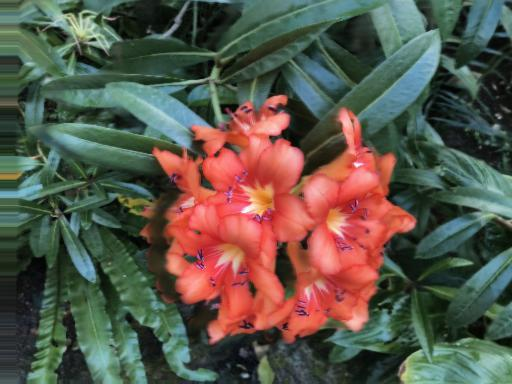}& 
     \includegraphics[width=0.16\textwidth]{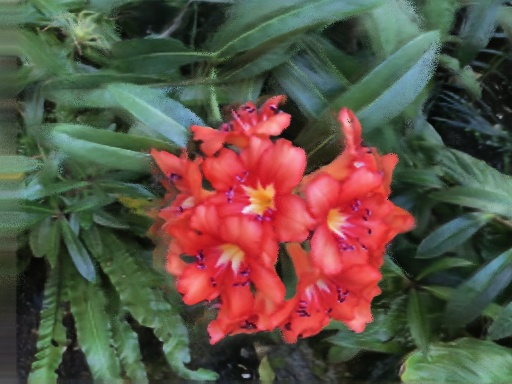}& 
     \includegraphics[width=0.16\textwidth]{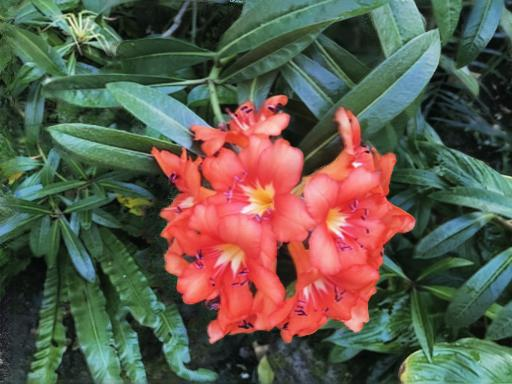}& \includegraphics[width=0.16\textwidth]{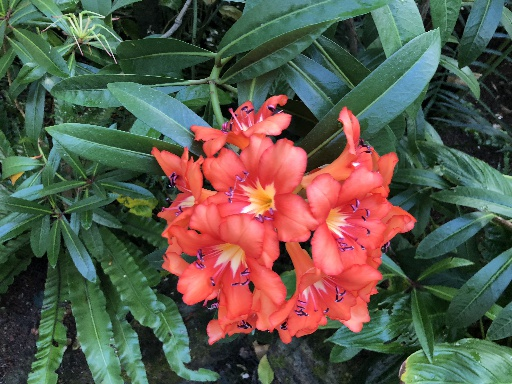} \\
     
\end{tabular}
}
    \caption{Visual comparisons on NeRF-LLFF scenes. Models are trained on the LLFF training set and evaluated on the LLFF test set.  Compared with  MPI, MINE and PixelNeRF, our DT-NeRF produces clear scenes with more details and fewer blurs or floats.} 
    % \vspace{-8pt}
    \label{fig:llff_rgb_supp}
\end{figure*}
\subsection{View Synthesis on NeRF-LLFF}

In table \ref{tab:llff_compare}, we compare our method with  PixelNeRF~\cite{Yu_2021_CVPR} and MINE ~\cite{Li_2021_ICCV2}  on the LLFF dataset.  We also compare to the pre-trained MPI~\cite{Tucker_2020_CVPR} on LLFF using the provided checkpoint. % 
Our method achieves the best performance in all the evaluation metrics of PSNR, SSIM and LPIPS. It outperforms the MINE  by 20\% in SSIM, 7\% in PSNR and 18\% in LPIPS.
\begin{table}[t]
\caption{Comparison with MPI, PixelNeRF and MINE on NeRF-LLFF dataset. }
% \vspace{1mm}
\centering
 \setlength{\abovecaptionskip}{4pt}
 \setlength{\belowcaptionskip}{-6pt}
% \small 
\begin{tabular}{c|ccc}
\hline 
Method     & LPIPS$\downarrow$  & SSIM$\uparrow$  & PSNR$\uparrow$ \\
\hline
MPI\cite{Tucker_2020_CVPR} & 0.502     & 0.356 & 14.6 \\ 
PixelNeRF \cite{Yu_2021_CVPR} & 0.476    & 0.468 & 17.5 \\
MINE \cite{Li_2021_ICCV2} & 0.386 & 0.531 & 18.2 \\
Ours& \textbf{0.317} & \textbf{0.637} & \textbf{19.5} \\
\hline
\end{tabular}
% \caption{Comparison with MPI, PixelNeRF and MINE on NeRF-LLFF dataset. }
\label{tab:llff_compare}
%\vspace{-4pt}
\end{table}
As shown in Figure \ref{fig:llff_rgb_supp}, our method is able to render novel views with higher quality (more fine details and fewer blurs).   Moreover, the depth rendering results of our DT-NeRF is also far better than MINE (Figure \ref{fig:llff_depth}).

Our method also far outperforms the PixelNeRF~\cite{Yu_2021_CVPR}.
As shown in Figure \ref{fig:llff pixelnerf},  we compare our DT-NeRF with PixelNeRF. We only use our volume rendering which just takes 16 samples in rendering the image view. As a comparison, PixelNeRF needs 96 samples.  With just $1/6$ point samples,  our volume rendering can synthesize more realistic images than PixelNeRF. This is because our volume rendering is supervised by the depth teacher and learns better 3D geometry. Moreover, the point sampling is guided by the coarse planar sampling results which assist the more
precise sampling around the object surfaces and avoids sampling a large number of useless points. 
\begin{table}[ht]
\caption{Ablation study on NeRF-LLFF on different rendering components.}
% \vspace{1mm}
% \small 
\centering
 \setlength{\abovecaptionskip}{4pt}
 \setlength{\belowcaptionskip}{-6pt}
\begin{tabular}{c|ccc}
\hline 
Method  & LPIPS$\downarrow$  & SSIM$\uparrow$  & PSNR$\uparrow$ \\
\hline
PixelNeRF \cite{Yu_2021_CVPR} & 0.476    & 0.468 & 17.5 \\
Planar rendering & 0.326 & 0.608 & 18.9 \\
%Plane+inpaint(w/o $\mathcal{L}'_d$) & 0.318 & 0.597 & 19.4 \\
Volume rendering & 0.351 & 0.628 & 18.8 \\
Planar+Volume & 0.338 & \textbf{0.637} & 19.0 \\
%Volume(16)+inpaint & 0.326 & 0.629 & 19.3 \\
Full settings & \textbf{0.317} & \textbf{0.637} & \textbf{19.5} \\
\hline
\end{tabular}
% \caption{Ablation study on NeRF-LLFF on different rendering components.}
\label{tab:baseline_compare_detail}
%\vspace{-8pt}
\end{table}

\begin{figure*}[t!] \small \centering
\setlength{\tabcolsep}{1pt}
\begin{tabular}{ccccc}
     Input  & MINE & Ours & Target GT\\ 
     \includegraphics[width=0.24\textwidth]{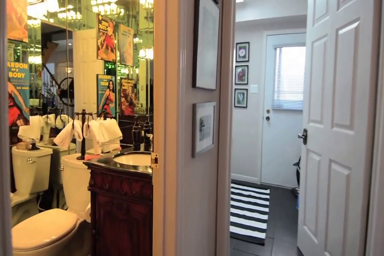}  &
     \includegraphics[width=0.24\textwidth]{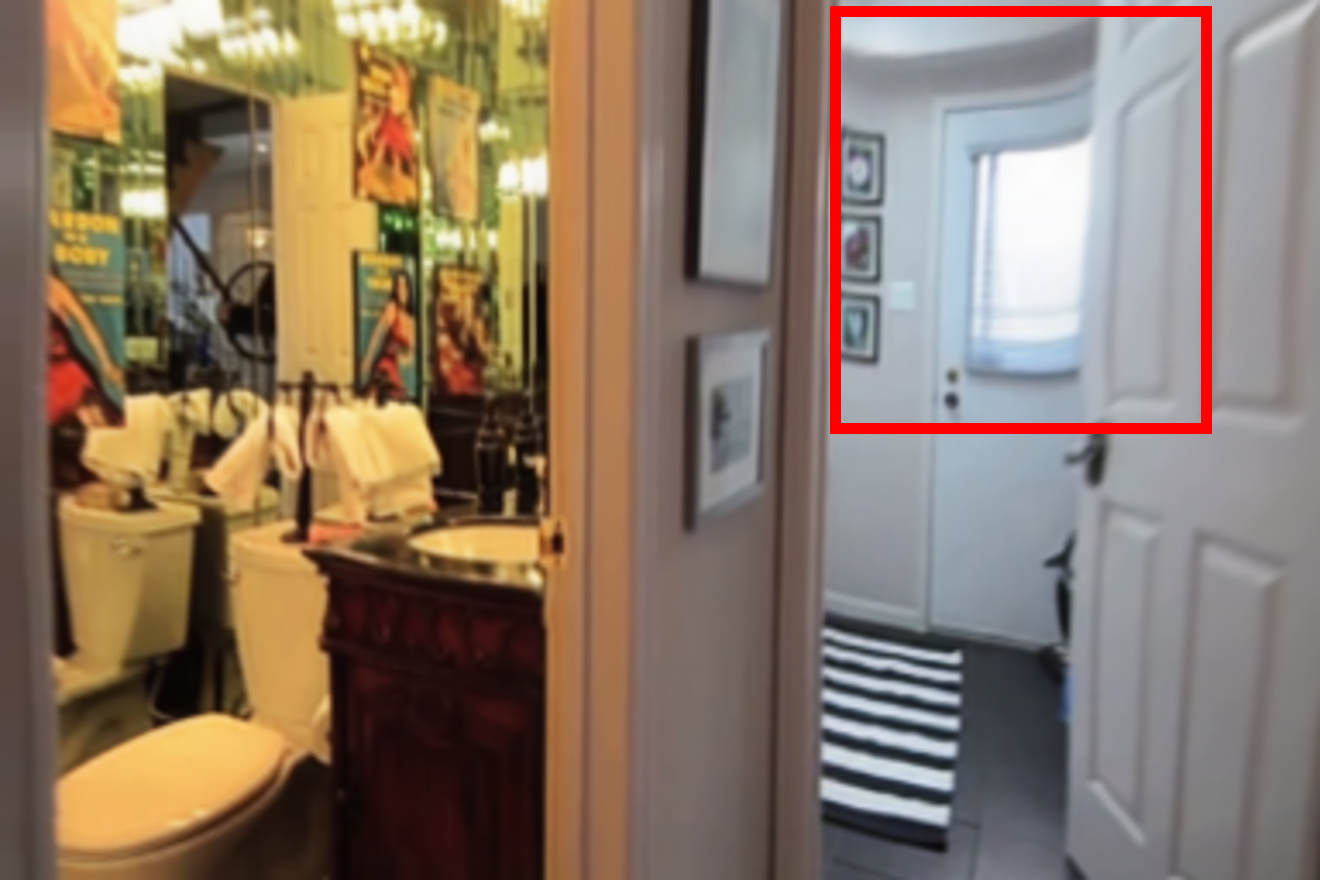} &
     \includegraphics[width=0.24\textwidth]{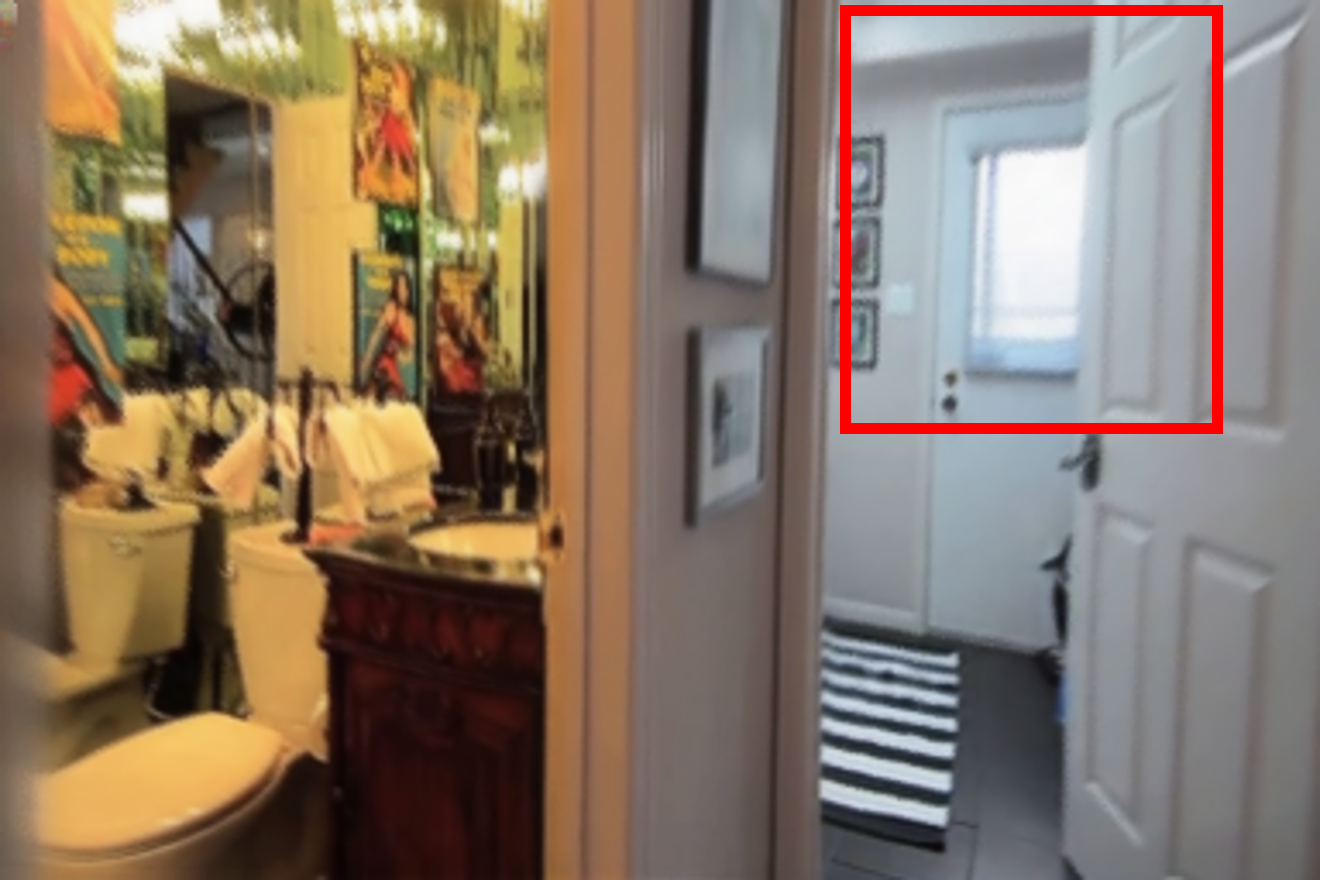}&
     \includegraphics[width=0.24\textwidth]{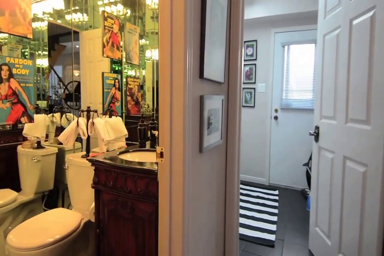} 
     \\
     \includegraphics[width=0.24\textwidth]{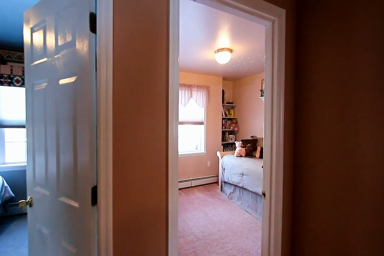}  &
     \includegraphics[width=0.24\textwidth]{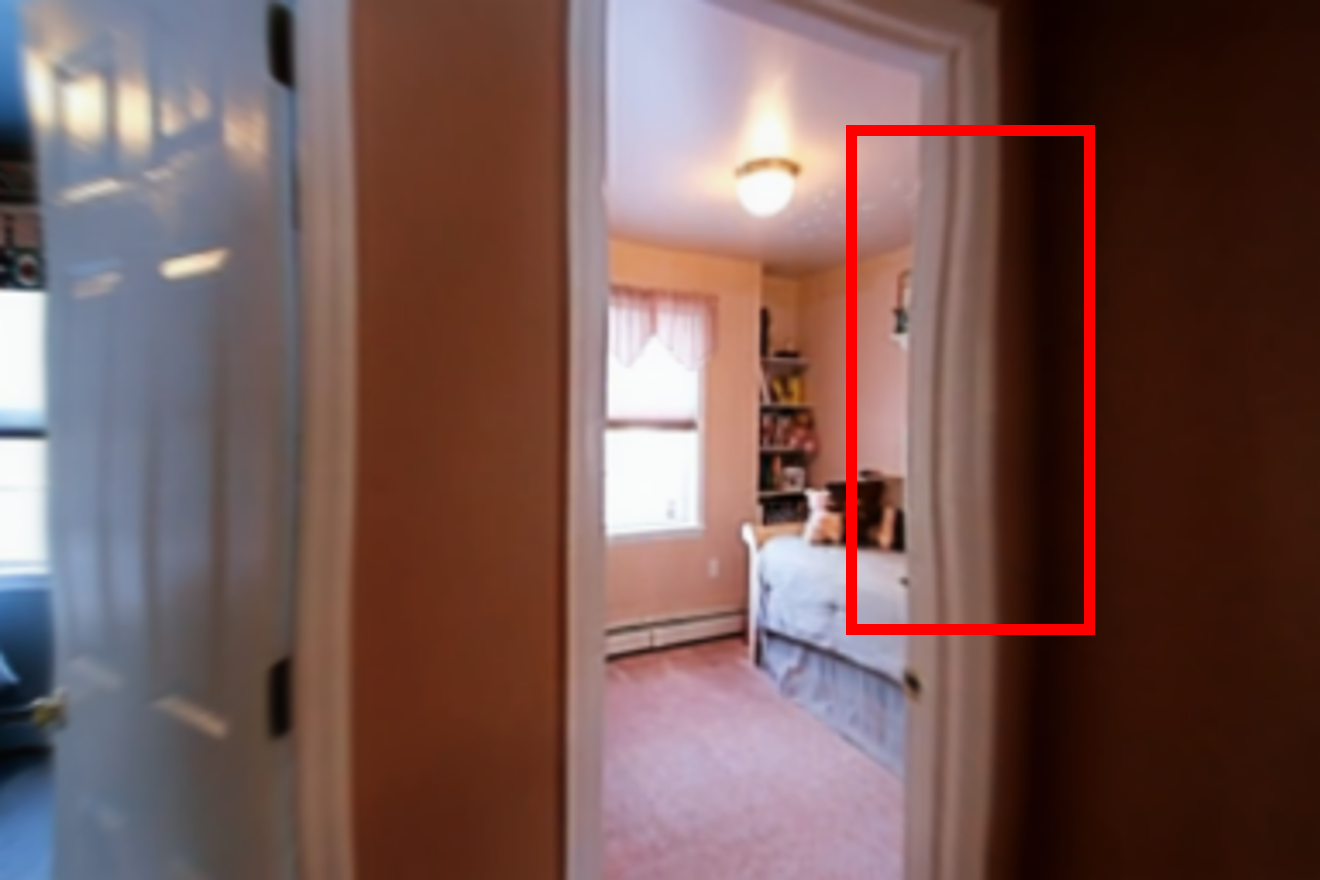} &
     \includegraphics[width=0.24\textwidth]{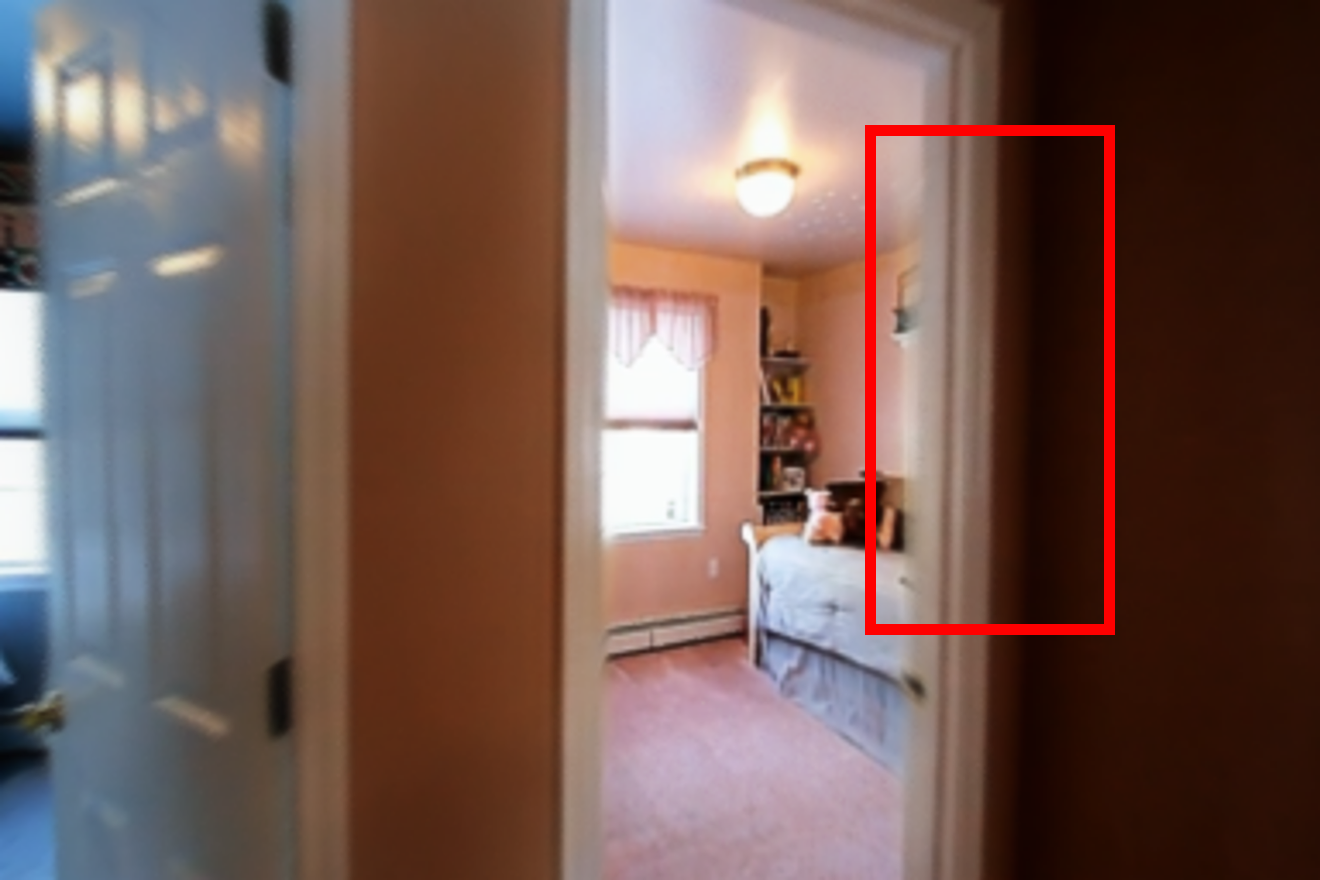}&
     \includegraphics[width=0.24\textwidth]{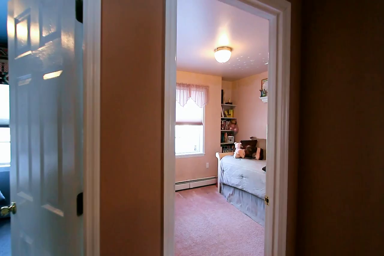}
     \\

     % \includegraphics[width=0.24\textwidth]{figure/realestate inpainting/0216_src.png} &
     % \includegraphics[width=0.24\textwidth]{figure/realestate/0216_MINE.png} &
     % \includegraphics[width=0.24\textwidth]{figure/realestate/0216_ours_f_inpaint.png}  &
     % \includegraphics[width=0.24\textwidth]{figure/realestate inpainting/0216_tgt.png}
     % \\
     % \includegraphics[width=0.24\textwidth]{figure/realestate inpainting/0228_src.png}  &
     % \includegraphics[width=0.24\textwidth]{figure/realestate/0228_MINE.png} &
     % \includegraphics[width=0.24\textwidth]{figure/realestate/0228_ours_f_inpaint.png} &
     % \includegraphics[width=0.24\textwidth]{figure/realestate inpainting/0228_tgt.png}
\end{tabular}
    \caption{Visual comparisons on RealEstate10K. Since our depth prediction is more accurate than MINE, our method can avoid many image distortions (as highlighted by the red windows).
    } 
    \label{fig_realestate_vis}
% \vspace{-4pt}
\end{figure*} 
\begin{table*}[ht]
\caption{Depth estimation results on NYU-Depth V2. $\uparrow$ denotes higher is better and $\downarrow$ denotes lower is better. By training on LLFF or a small part of RealEstate10K dataset, our DT-NeRF far outperforms both MPI and MINE that are trained on the same dataset. 
}
\centering
\small
\resizebox{0.95\textwidth}{!}{
\begin{tabular}{ccc | cccccc }
\hline
    &    &   & \multicolumn{6}{c}{NYU-Depth V2 \cite{silberman2012indoor}}  \\ 
Method        & Supervision & Dataset            & rel$\downarrow$           & log10$\downarrow$         &
RMS$\downarrow$          & $\sigma$1$\uparrow$        & $\sigma$2$\uparrow$       & $\sigma$3$\uparrow$  \\
\hline
MPI \cite{Tucker_2020_CVPR}        & RGB   & RealEstate10K      & 0.18          & 0.07         & 0.60         & 0.74          & 0.94          & 0.98        \\
MINE  \cite{Li_2021_ICCV2} & RGB   & LLFF      & 0.32          & 0.12          & 0.93         & 0.51          & 0.81          & 0.92   \\
MINE  \cite{Li_2021_ICCV2} & RGB   & RealEstate10K(small)      & 0.18          & 0.07          & 0.58         & 0.75          & 0.94          & 0.98   \\
Ours  & RGB   & LLFF      &   0.14        &    0.06       &  0.48  &   0.82       &     0.97      &  0.99  \\
Ours & RGB   & RealEstate10K(small)      &   \textbf{0.12}        &   \textbf{0.05}       &  \textbf{0.43}  &   \textbf{0.86}       &    \textbf{0.97}      &  \textbf{0.99}  \\
%ours new ($N = 32$) & RGB   & RealEstate10K(small)      &   0.17        &    0.07       &  0.56  &   0.76       &     0.94      &  0.98  \\
\hline
\end{tabular}}
\label{table_depth_estimation}
%\vspace{-2pt}
\end{table*}
\subsection{View Synthesis on RealEstate10K}\label{sec:realestate}
We also evaluate our method on the RealEstate10K dataset. It is compared with the state-of-the-art MINE. Our method achieves the best PSNR, SSIM and LPIPS. It outperforms the MPI~\cite{Tucker_2020_CVPR} by 4$\sim$10\% in these three evaluation metrics, and also surpass the MINE~\cite{Li_2021_ICCV2} with a better RGB rendering quality. More importantly, compared with MPI and MINE, our DT-NeRF produces far better depth predictions with sharper depth edges and more accurate estimations in occluded regions in Figure \ref{fig_realestate_vis}.

\begin{table}[t]
\caption{Evaluation on RealEstate10K. Models are trained on 1000 Realestate10k scenes and tested on 600 scenes.}

\centering
\begin{tabular}{c|ccc}
\hline 
Method     & LPIPS$\downarrow$  & SSIM$\uparrow$  & PSNR$\uparrow$ \\
\hline
MPI &   0.159   & 0.793 & 23.9 \\ 
MINE & 0.146 & 0.821 & 24.6 \\
Ours($N_f = 16$)& 0.152 & 0.831 & \textbf{25.0} \\
Ours($N_f = 32$)& \textbf{0.144} & \textbf{0.833} & 24.9 \\

\hline
\end{tabular}

\label{tab:realestate_compare}
\end{table}

\subsection{Depth Estimation on NYU-V2}
\begin{figure*}[t!] \centering
\setlength{\tabcolsep}{1.5pt}
\begin{tabular}{ccccc}
     Input  & MINE & Ours & GT \\ 
     \includegraphics[width=0.24\textwidth]{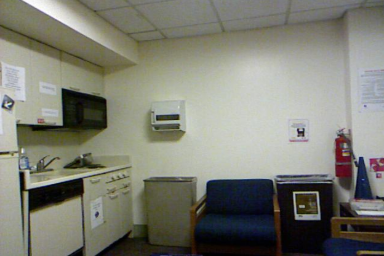}  & 
     \includegraphics[width=0.24\textwidth]{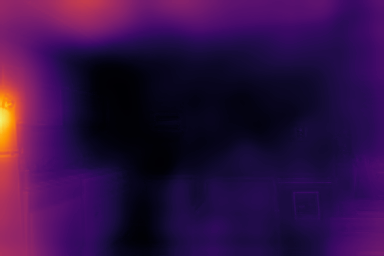} &
     \includegraphics[width=0.24\textwidth]{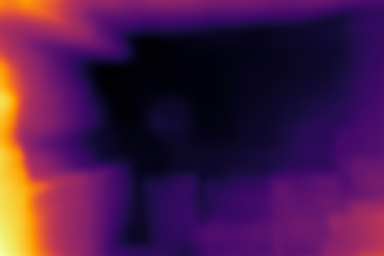} & 
     \includegraphics[width=0.24\textwidth]{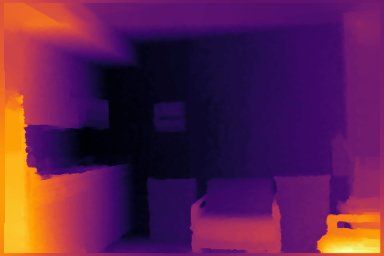}
     \\
     % \includegraphics[width=0.24\textwidth]{figure/nyu/0015_rgb.png} &
     % \includegraphics[width=0.24\textwidth]{figure/nyu/0015_MINE.png} &
     % \includegraphics[width=0.24\textwidth]{figure/nyu/0015_ours.png} &
     % \includegraphics[width=0.24\textwidth]{figure/nyu/0015_gt.png}
     % \\
     % \includegraphics[width=0.24\textwidth]{figure/nyu/0003_rgb.png}  &
     % \includegraphics[width=0.24\textwidth]{figure/nyu/0003_MINE.png} &
     % \includegraphics[width=0.24\textwidth]{figure/nyu/0003_ours.png} &
     % \includegraphics[width=0.24\textwidth]{figure/nyu/0003_gt.png}
     % \\
     % \includegraphics[width=0.24\textwidth]{figure/nyu/0017_rgb.png}  &
     % \includegraphics[width=0.24\textwidth]{figure/nyu/0017_MINE.png} &
     % \includegraphics[width=0.24\textwidth]{figure/nyu/0017_ours.png} &
     % \includegraphics[width=0.24\textwidth]{figure/nyu/0017_gt.png}
     % \\
     \includegraphics[width=0.24\textwidth]{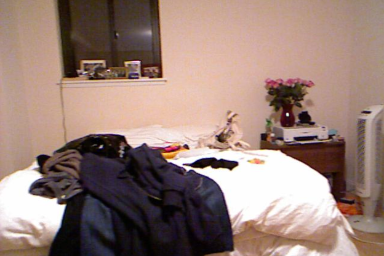} &
     \includegraphics[width=0.24\textwidth]{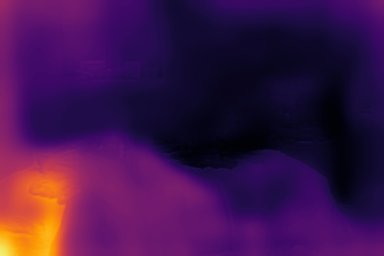} &
     \includegraphics[width=0.24\textwidth]{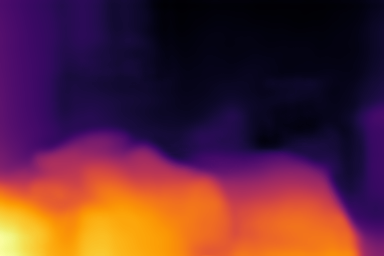}  &
     \includegraphics[width=0.24\textwidth]{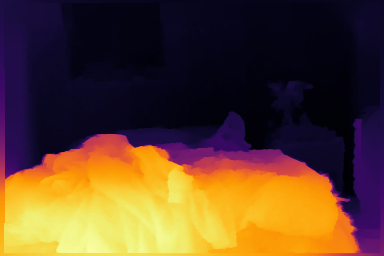}
     \\
\end{tabular}
    \caption{Qualities of depth maps on NYU-Depth V2 dataset. Models are trained on the RealEstate10K dataset. Our method produces better depth maps with more structure details and sharper depth edges.} 
    % \vspace{-8pt}
    \label{fig_depth_vis_nyu}
\end{figure*} 
We evaluate depth estimation on NYU-Depth V2 dataset~\cite{silberman2012indoor}. We perform our method on the labeled subset which consists of 1449 densely labeled pairs of RGB and depth images taken from a variety of indoor scenes. To solve the scale ambiguity problem of predicted depth, following \cite{niklaus20193d,Li_2021_ICCV2}, we scale and bias the predicted depth to minimize the $L2$ depth error with respect to ground truth. 

As shown in Table \ref{table_depth_estimation}, our model performs far better than MINE and MPI in all the evaluation metrics. Even with our model trained on LLFF which consists of only 8 scenes, it still has better depth results than MINE trained on the larger RealEstate10k dataset (1000 scenes and 32,000 training pairs). 
This is because our DT-NeRF training is guided by the dense pseudo depth maps which help it learn consistent 3D geometry across the source views and the target views in both the planar rendering and the volume rendering. Some results are shown in Figure \ref{fig_depth_vis_nyu}.

 \begin{table}[t]

\caption{Generalization abilities on unseen ``room'' in LLFF dataset. Note that SinNeRF and PixelNeRF are fine-tuned on the test scenes.}
% \small 
\centering
\begin{tabular}{cc|ccc}
\hline 
Method & Dataset & LPIPS$\downarrow$  & SSIM$\uparrow$  & PSNR$\uparrow$ \\
\hline
SinNeRF & LLFF-room  &  0.431   &  0.642  &  18.10  \\
PixelNeRF & LLFF & 0.419    &  0.675    &  18.23  \\
Ours & Realestate10k & 0.488   &  0.613   & 18.09  \\
Ours & LLFF & 0.339 & 0.745 & 20.65 \\
\hline
\end{tabular}
\label{tab:sinnerf}
% \vspace{-8pt}
\end{table}
 \begin{figure*}[t!] \centering
\setlength{\tabcolsep}{1.5pt}
\begin{tabular}{ccccc}
     Input  & Teacher & Student(w/o $\mathcal{L}'_d$)&Student& GT \\ 
     \includegraphics[width=0.19\textwidth]{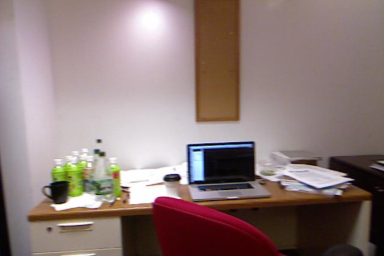}  & 
     \includegraphics[width=0.19\textwidth]{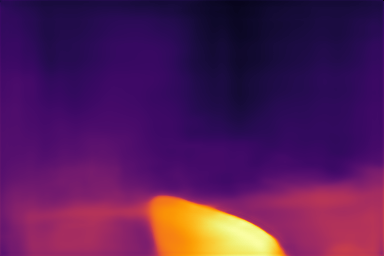} &
     \includegraphics[width=0.19\textwidth]{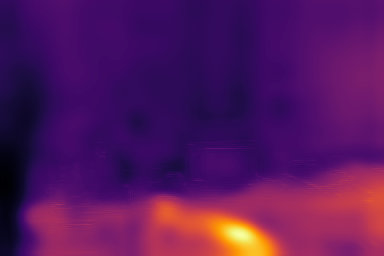} &
     \includegraphics[width=0.19\textwidth]{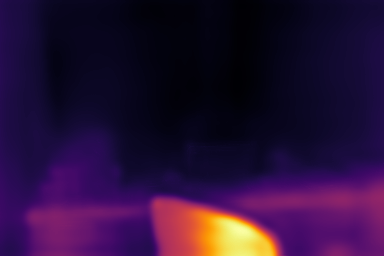}& 
     \includegraphics[width=0.19\textwidth]{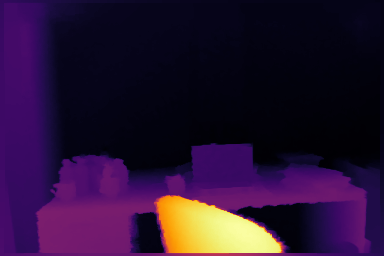}
     \\
     \includegraphics[width=0.19\textwidth]{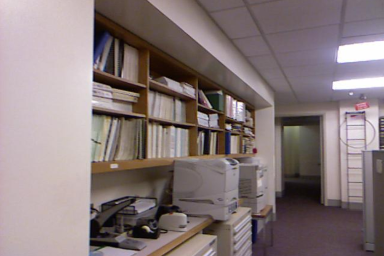} & 
     \includegraphics[width=0.19\textwidth]{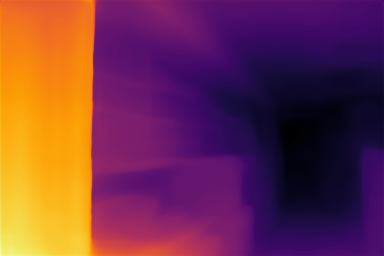} &
     \includegraphics[width=0.19\textwidth]{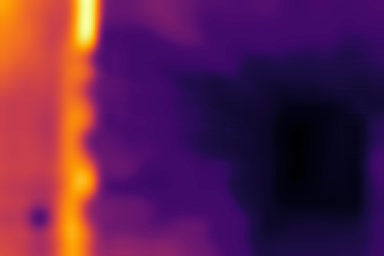} &
     \includegraphics[width=0.19\textwidth]{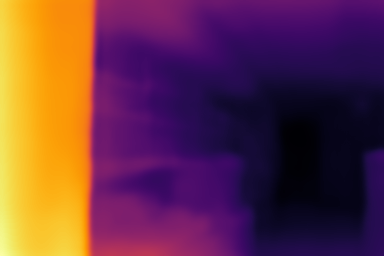} & 
     \includegraphics[width=0.19\textwidth]{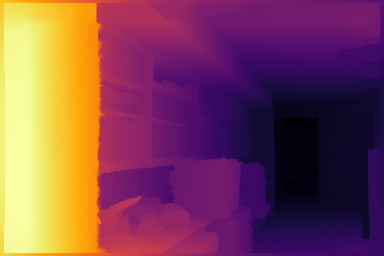}
     \\
     \includegraphics[width=0.19\textwidth]{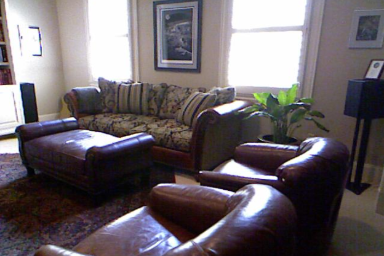} & 
     \includegraphics[width=0.19\textwidth]{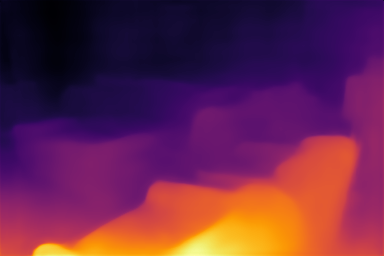} &
     \includegraphics[width=0.19\textwidth]{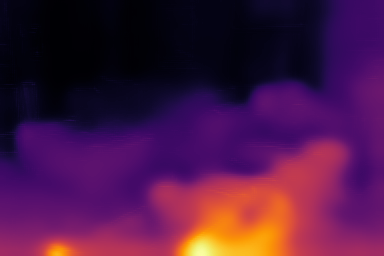} &
     \includegraphics[width=0.19\textwidth]{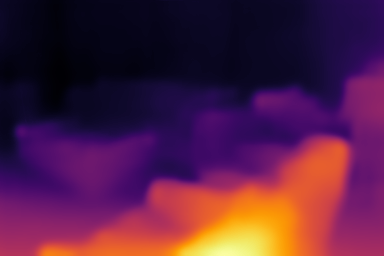} & 
     \includegraphics[width=0.19\textwidth]{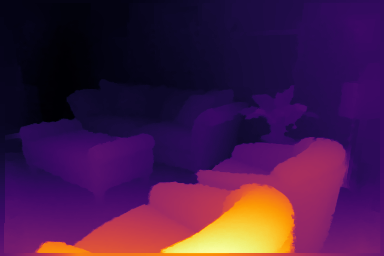} \\
\end{tabular}
    \caption{Comparisons between the teacher net and the student net. Model is trained RealEstate10K. The predicted depth results by the student net and the teacher net are tested on  NYU-depth V2 respectively. The depth supervision with pseudo depth loss $\mathcal{L}'_d$ can significantly improve the quality of the student net's prediction. } 
    % \vspace{-8pt}
    \label{fig stu tea}
\end{figure*} 
\begin{table*}[h]
% \small 
\centering
\caption{Student/teacher net trained/fine-tuned on RealEstate10K (small) and  evaluated on NYU-depth V2.}
\begin{tabular}{c | cccccc }
\hline
Method & rel$\downarrow$  & log10$\downarrow$  &
RMS$\downarrow$    & $\sigma$1$\uparrow$  &$\sigma$2$\uparrow$ & $\sigma$3$\uparrow$  \\
\hline
Teacher & 0.127   & 0.054 & 0.452  & 0.853 & \textbf{0.973}  &  \textbf{0.993}  \\
Student(w/o $\mathcal{L}'_d$) & 0.138   & 0.058 & 0.487  & 0.830 & 0.964  &  0.990  \\
Student & \textbf{0.123}        &   \textbf{0.052}       &  \textbf{0.434}  &   \textbf{0.859}       &    0.972      &  0.992 \\
\hline
\end{tabular}
\label{tab:stu_tea}
\end{table*}
\subsection{Analysis on Student and Teacher Net}
Table \ref{tab:stu_tea} shows $\mathcal{L}'_d$ will improve student net's prediction quality of depth map. And the student net even has stronger generalization performance than fine-tuned teacher net on the depth prediction. 
In Figure \ref{fig stu tea}, we show comparisons of the predicted depth maps between the depth teacher and the student net. We use the models trained on RealEstate10K(small) to inference depth maps on the NYUv2 dataset. Student net performs a little better in four  of the six evaluation metrics.

Although teacher net (monocular depth estimation) is able to predict high-quality depth maps on the input image, it can't render depth maps for novel view. The  student net renders both the RGB and the depth maps for the novel views which are very important for applications like video editing, augmented reality etc.
 
\subsection{Generalization}
Our DT-NeRF is shown able to generalize to new scenes without fine-tuning or retraining on each of them. This is because we design a depth teacher net to supervise the
joint student rendering mechanism and boost the learning of consistent 3D geometry. In this section, we  pretrain our DT-NeRF on the RealEstate10K datasets and then compare it with the fine-tuned (on the target NeRF-LLFF scenes) SinNeRF and PixelNeRF on the unseen NeRF-LLFF scenes. 
\paragraph{Comparison with Fine-tuned SinNeRF}
Our DT-NeRF shares a similarity with SinNeRF~\cite{sinnerf}. Both models use a pre-trained teacher net to supervise the training process, while DT-NeRF uses a dense monocular depth teacher, SinNeRF uses a semantic teacher and a geometry teacher. 
SinNeRF has no generalization abilities to unseen data/images. It needs to be retrained or fine-tuned on each scene. We train SinNeRF for each scene (512$\times$384) and use the ground truth depth \cite{sinnerf} to supervise. %The training on each scenes takes 24 hours.   
Our method has excellent generalization abilities to new scenes. So there is no need to fine-tune it on the test scenes. 
We directly use our model trained on a small part of RealEstate10K dataset (1000 scenes) and perform novel view synthesis on ``room'' scene in LLFF dataset. After training on the ``room'' scene, SinNeRF achieves 18.10 in PSNR. Our unadapted method achieves a similar 18.09 in PSNR.

\paragraph{Comparison with Fine-tuned PixelNeRF}
We also compare our DT-NeRF (trained on unrelated RealEstate10K dataset) with the PixelNeRF~\cite{Yu_2021_CVPR} that is trained/fine-tuned on the LLFF dataset (including the test room scenes). Even without seeing the LLFF scenes during the training,  our method can still produce a similar performance (18.09 in PNSR) that is close to the test-data-fine-tuned PixelNeRF (18.23 in PSNR). After fine-tuning, our DT-NeRF achieves a 14\% improvement in PSNR which is far better than the fine-tuned PixelNeRF and SinNeRF.

\section{Limitations and Failure Case}
Similar to all existing methods for single-image-based novel view synthesis, our method also fails when there are large changes or translations of view angles (e.g. $>60^{\circ}$). In this case, almost the whole novel-view image is in the occlusion part. A single-image input doesn't provide enough information for predicting such novel views and thus will produce many artifacts.    

\section{Conclusion}

In this paper, we have proposed DT-NeRF for photorealistic novel view rendering using only a single view image as input.
This is achieved by combining plane rendering and volume rendering for better rendering quality and better generalizations to new scenes. 
 We also design an effective depth teacher network that produces dense pseudo depths to supervise the joint rendering and learn consistent 3D geometries. 
 DT-NeRF is shown able to outperform state-of-the-art single-view NeRFs in both the RGB and the depth rendering. 

\bibliographystyle{IEEEtran}
\bibliography{ref.bib}
\end{document}